\documentclass{article}

\PassOptionsToPackage{numbers, compress, sort}{natbib}

\usepackage[final]{neurips_2021}

\usepackage[utf8]{inputenc} 
\usepackage[T1]{fontenc}    
\usepackage[pagebackref=true, colorlinks=true,  citecolor=blue]{hyperref}       
\usepackage{url}            
\usepackage{booktabs}       
\usepackage{amsfonts}       
\usepackage{nicefrac}       
\usepackage{microtype}      
\usepackage{xcolor,colortbl}         
\usepackage{amssymb}
\usepackage{amsmath,amsfonts}
\usepackage{amsopn}
\usepackage{bm} 
\usepackage{multirow}
\usepackage{flushend}
\usepackage{tabularx}
\newcommand{\vct}[1]{\boldsymbol{#1}} 

\newcommand{\ProbOpr}[1]{\mathbb{#1}}

\newcommand{\expect}[2]{%
\ifthenelse{\equal{#2}{}}{\ProbOpr{E}_{#1}}
{\ifthenelse{\equal{#1}{}}{\ProbOpr{E}\left[#2\right]}{\ProbOpr{E}_{#1}\left[#2\right]}}} 
\newcommand{\var}[2]{%
\ifthenelse{\equal{#2}{}}{\ProbOpr{VAR}_{#1}}
{\ifthenelse{\equal{#1}{}}{\ProbOpr{VAR}\left[#2\right]}{\ProbOpr{VAR}_{#1}\left[#2\right]}}} 

\newcommand{\vtheta}{\vct{\theta}}

\newcommand{\vx}{{\vct{x}}}
\newcommand{\vy}{\vct{y}}

\newcommand{\vw}{\vct{w}}

\newcommand{\eat}[1]{}

\usepackage{xspace}
\usepackage{wrapfig}
\usepackage{caption}
\usepackage{amssymb}
\usepackage{pifont}
\usepackage{draftfigure} 
\usepackage{makecell}
\usepackage{enumitem}

\newcommand{\eg}{{\em e.g.}}
\newcommand{\ie}{{\em i.e.}}

\newcommand{\cmark}{\ding{51}}%
\newcommand{\xmark}{\ding{55}}%
\newcommand\mypara[1]{\vspace{1.0mm}\noindent\textbf{#1}}
\newcommand{\ourmethod}{{\sc {NorCal}}\xspace}
\newcommand{\ourmethodbold}{{\textbf{\textsc {NorCal}}}\xspace}
\newcommand{\nor}{{\sc {Nor}}\xspace}
\newcommand{\cali}{{\sc {Cal}}\xspace}
\newcommand{\AP}[1]{${\text{AP}_{\textit{#1}}}$\xspace}
\newcommand{\APb}{${\text{AP}^{\textit{b}}}$\xspace}

\newcommand{\newadd}[1]{{\color{black}  #1}\xspace}

\definecolor{codegreen}{rgb}{0,0.6,0}
\definecolor{codegray}{rgb}{0.5,0.5,0.5}
\definecolor{codepurple}{rgb}{0.58,0,0.82}
\definecolor{backcolour}{rgb}{0.95,0.95,0.92}
\definecolor{darkgreen}{rgb}{0,0.4,0}
\definecolor{cerise}{rgb}{0.871, 0.192, 0.388}
\definecolor{carmine}{rgb}{0.59, 0.0, 0.09}
\definecolor{olive}{rgb}{0.332, 0.418, 0.184}
\definecolor{navyblue}{rgb}{0.496, 0.810, 0.837}

\definecolor{f_other}{rgb}{0.55,0.34,0.29}
\definecolor{c_other}{rgb}{0.58,0.40,0.74}
\definecolor{r_other}{rgb}{0.84,0.15,0.16}
\definecolor{rare}{rgb}{0.17,0.63,0.17}
\definecolor{my_a}{rgb}{1.0, 0.49, 0.06}
\definecolor{my_b}{rgb}{0.12, 0.47, 0.70}
\definecolor{my_c}{rgb}{0.17, 0.62, 0.17}

\definecolor{LightCyan}{rgb}{0.88,1,1}
\definecolor{Gray}{gray}{0.45}

\title{On Model Calibration for Long-Tailed \\Object Detection and Instance Segmentation}

\author{Tai-Yu Pan$^{1}$\thanks{Equal contributions} \quad Cheng Zhang$^1$\footnotemark[1] \quad Yandong Li$^2$ \quad Hexiang Hu$^2$\\
\textbf{Dong Xuan$^1$ \quad Soravit Changpinyo$^2$ \quad Boqing Gong$^2$ \quad Wei-Lun Chao$^1$} \\[5pt]
{$^1$The Ohio State University \quad  $^2$Google Research}
}

\begin{document}

\maketitle


\begin{abstract}
Vanilla models for object detection and instance segmentation suffer from the heavy bias toward detecting frequent objects in the long-tailed setting.
Existing methods address this issue mostly during training, \eg, by re-sampling or re-weighting. In this paper, we investigate a largely overlooked approach --- \emph{post-processing calibration} of confidence scores. We propose \ourmethod, \textbf{Normalized Calibration} for long-tailed object detection and instance segmentation, a simple and straightforward recipe that reweighs the predicted scores of each class by its training sample size. We show that separately handling the background class and normalizing the scores over classes for each proposal are keys to achieving superior performance. On the LVIS dataset, \ourmethod
can effectively improve nearly all the baseline models not only on rare classes but also on common and frequent classes.  Finally, we conduct extensive analysis and ablation studies to offer insights into various modeling choices and mechanisms of our approach.
\newadd{Our code is publicly available at~\url{https://github.com/tydpan/NorCal}}.
\end{abstract}

\section{Introduction}
\label{s_intro}

Object detection and instance segmentation are the fundamental tasks in computer vision and have been approached from various perspectives over the past few decades~\cite{li2014secrets,gupta2014learning,viola2001robust,papageorgiou1998general,felzenszwalb2009object}. With the recent advances in neural networks~\cite{he2017mask,ren2016faster,girshick2014rich,liu2016ssd,redmon2016you,bochkovskiy2020yolov4,redmon2017yolo9000,peng2020deep,lin2017focal,lin2017feature,chen2018masklab}, we have witnessed an unprecedented breakthrough in detecting and segmenting frequently seen objects such as people, cars, and TVs~\cite{han2018advanced,hafiz2020survey,zou2019object,liu2020deep,jiao2019survey}. Yet, when it comes to detect rare, less commonly seen objects (\eg, walruses, pitchforks, seaplanes, etc.)~\cite{gupta2019lvis,inat}, there is a drastic performance drop largely due to insufficient training samples~\cite{tan20201st,zhoujoint}. How to overcome the ``long-tailed'' distribution of different object classes~\cite{zhu2014capturing} has therefore attracted increasing attention lately~\cite{tan2020equalization,wang2020devil,li2020overcoming,oksuz2020imbalance}.

To date, most existing works tackle this problem in the \emph{model training phase}, \eg, by developing algorithms, objectives, or model architectures to tackle the long-tailed distribution~\cite{hu2020learning,li2020overcoming,wang2020frustratingly,tan2020equalization,wu2020forest,gupta2019lvis,wang2020seesaw,tan2020equalizationv2,wang2020devil}. \citet{wang2020devil} investigated the widely used instance segmentation model Mask R-CNN~\cite{he2017mask} and found that the performance drop comes primarily from \emph{mis-classification of object proposals}. Concretely, the model tends to give frequent classes higher confidence scores~\cite{dave2021evaluating}, hence biasing the label assignment towards frequent classes. This observation suggests that 
techniques of class-imbalanced learning~\cite{CBfocal,buda2018systematic,he2009learning,ren2020balanced} can be applied to long-tailed detection and segmentation.

Building upon the aforementioned observation, we take another route in the \emph{model inference phase} by explicit \emph{post-processing  calibration}~\cite{buda2018systematic,menon2020long,kang2019decoupling,ye2020identifying,kim2020adjusting}, which adjusts a classifier's confidence scores among classes, without changing its internal weights or architectures. Post-processing calibration is efficient and widely applicable since it requires no re-training of the classifier. Its effectiveness on multiple imbalanced classification benchmarks~\cite{wu2021adversarial,kang2019decoupling} may also translate to long-tailed object detection and instance segmentation.

\begin{figure}[t]
    \centering
    \includegraphics[width=1\textwidth]{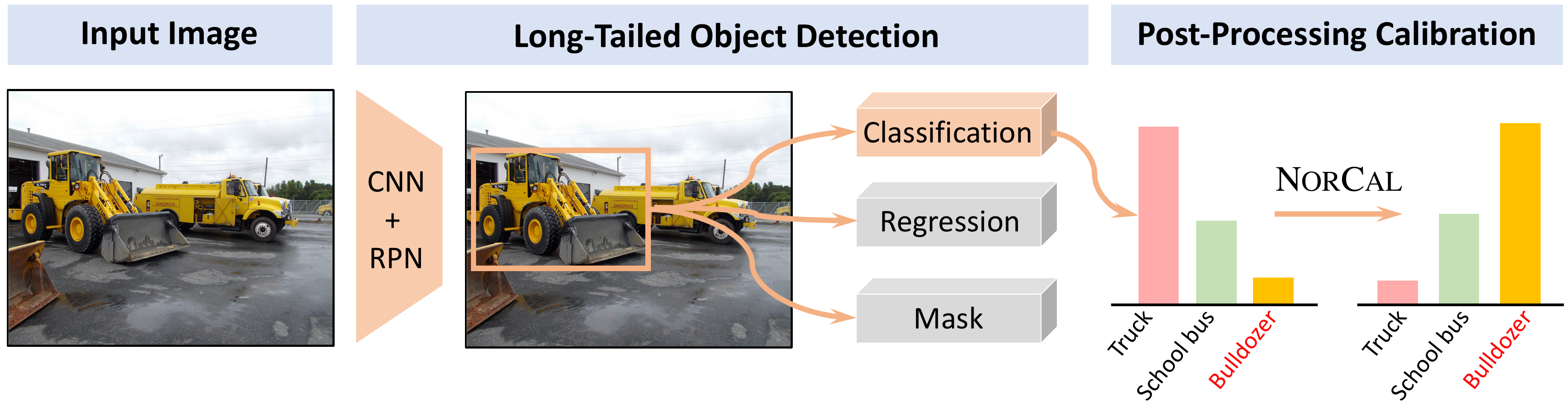}
    \vskip -5pt
    \caption{\small \textbf{Normalized Calibration (\ourmethodbold).} Object detection or instance segmentation models (\eg, \cite{ren2016faster,he2017mask}) trained with data from a long-tailed distribution tend to output higher confidence scores for the head classes (\eg, ``Truck'') than for the tail ones (\eg, the true class label ``{\color{red}{Bulldozer}}''). \ourmethod investigates a simple but largely overlooked approach to correct this mistake --- post-processing calibration of the classification scores \emph{after training} --- and significantly improves nearly all the models we consider.}
    \label{fig:overview}
    \vskip -10pt
\end{figure}

{In this paper, we propose a simple post-processing calibration technique inspired by class-imbalanced learning~\cite{ye2020identifying,menon2020long} and show that it can significantly improve a pre-trained object detector's performance on detecting both rare and common classes of objects.} \emph{We note that our results are in sharp contrast to a couple of previous attempts on exploring post-processing calibration in object detection~\cite{li2020overcoming,dave2021evaluating}, which reported poor performance and/or sensitivity to hyper-parameter tuning.} We also note that the calibration techniques in \cite{wang2019classification,wang2020devil} are implemented in the training phase and are not post-processing.

Concretely, we apply post-processing calibration to the classification sub-network of a pre-trained object detector. Taking Faster R-CNN~\cite{ren2016faster} and Mask R-CNN~\cite{he2017mask} for examples, they apply to each object proposal a $(C+1)$-way softmax classifier, where $C$ is the number of foreground classes, and 1 is the background class. To prevent the scores from being biased toward frequent classes~\cite{wang2020devil,dave2021evaluating},
we \emph{re-scale the logit} of every class according to its class size, \eg, number of training images. 
Importantly, we leave the logit of the background class intact because (a) the background class has a drastically different meaning from object classes and 
(b) its value does not affect the ranking among different foreground classes.
After adjusting the logits, we then re-compute the confidence scores (with normalization across all classes, including the background) to decide the label assignment for each object proposal\footnote{Popular evaluation protocols allow multiple labels per proposal if their confidence scores are high enough.} (see \autoref{fig:overview}). We note that it is crucial to normalize the scores across all classes
since it triggers \emph{re-ranking of the detection results within each class} (see \autoref{fig:norm}), influencing the class-wise precision and recall.
Instead of separately adjusting each class by a specific factor~\cite{dave2021evaluating}, we follow~\cite{menon2020long,ye2020identifying,CBfocal} to set the factor as a function of the class size, leaving only one hyper-parameter to tune. We find that it is robust to use the training set to set this hyper-parameter,  making our approach applicable to scenarios where collecting a held-out representative validation set is challenging. 

Our approach, named \textbf{Normalized Calibration} for long-tailed object detection and instance segmentation (\ourmethod), is model-agnostic as long as the detector has a softmax classifier or multiple binary sigmoid classifiers for the objects and the background. We validate \ourmethod on the LVIS~\cite{gupta2019lvis} dataset for both long-tailed object detection and instance segmentation. \ourmethod can consistently improve not only baseline models (\eg, Faster R-CNN~\cite{ren2016faster} or Mask R-CNN~\cite{he2017mask}) but also many models that are dedicated to the long-tailed distribution.
Hence, our best results notably advance the state of the art. Moreover, \ourmethod can improve both the standard average precision (AP) and the category-independent AP$^\text{Fixed}$ metric~\cite{dave2021evaluating}, implying that \ourmethod does not trade frequent class predictions for rare classes but rather \emph{improve the proposal ranking within each class}. Indeed, through a detailed analysis, we show that \ourmethod can in general improve both the precision and recall for each class, making it appealing to almost any existing evaluation metrics. Overall, we view \ourmethod a simple plug-and-play component to improve object detectors' performance during inference. 

\section{Related Work}
\label{s_related}

\mypara{Long-tailed detection and segmentation.} Existing works on long-tailed object detection can roughly be categorized into {re-sampling}, {cost-sensitive learning}, and {data augmentation}. \emph{Re-sampling methods} change the long-tailed training distribution into a more balanced one by sampling data from rare classes more often~\cite{gupta2019lvis,chang2021image,shen2016relay}. \emph{Cost-sensitive learning} aims at adjusting the loss of data instances according to their labels~\cite{tan2020equalization,tan2020equalizationv2,hsieh2021droploss,wang2020seesaw}. Building upon these, some methods perform two- or multi-staged training~\cite{wang2020frustratingly,wang2020devil,kang2019decoupling,hu2020learning,ren2020balanced,li2020overcoming,zhang2021distribution,wang2021adaptive}, which first pre-train the models in a conventional way, using data from all or just the head classes; the models are then fine-tuned on the entire long-tailed data using either re-sampling or cost-sensitive learning. 
Besides, another thread of works leverages {data augmentation} for the object instances of the tail classes to improve long-tailed object detection~\cite{ghiasi2020simple,ramanathan2020dlwl,zhang2021simple,zang2021fasa}. 

In contrast to all these previous works, we investigate post-processing calibration~\cite{kang2019decoupling,ye2020identifying,kim2020adjusting,tian2020posterior,menon2020long} to adjust the learned model in the testing phase, without modifying the training phase or modeling. Concretely, these methods adjust the predicted confident scores (\ie, the posterior over classes) for each test instance, \eg, by normalizing the classifier norms~\cite{kang2019decoupling} or by scaling or reducing the logits according to class sizes \cite{ye2020identifying,menon2020long,kim2020adjusting}. Post-processing calibration is quite popular in imbalanced classification but not in long-tailed object detection. 
To our knowledge, only \citet{li2020overcoming} and \citet{tang2020long} have studied this approach for object detection\footnote{Calibration in \cite{wang2019classification,wang2020devil} is in the training phase and is not post-processing. \iffalse We compare to them in \autoref{tab:suppl_v0.5_seg}.\fi} . \citet{li2020overcoming} applied classifier normalization~\cite{kang2019decoupling} as a baseline but showed inferior results; \citet{tang2020long} developed causal inference calibration rules, which however require a corresponding de-confounded training step. \citet{dave2021evaluating} applied methods for calibrating model uncertainty, which are quite different from class-imbalanced learning (see the next paragraph).
In this paper, we demonstrate that existing calibration rules for class-imbalanced learning~\cite{ye2020identifying,menon2020long,kim2020adjusting} can significantly improve long-tailed object detection, if paired with appropriate ways to deal with the background class and normalized the adjusted logits. 
We refer the reader to the {\color{magenta}{supplementary material}} for a comprehensive survey and comparison of the literature.

\mypara{Calibration of model uncertainty.} The calibration techniques we employ are different from the ones used for calibrating model uncertainty~\cite{guo2017calibration,platt1999probabilistic,zadrozny2001obtaining,naeini2015obtaining,zadrozny2002transforming,kull2017beta,kull2019beyond}: we aim to adjust the prediction across classes, while the latter adjusts the predicted probability to reflect the true correctness likelihood. 
Specifically for long-tailed object detection, \citet{dave2021evaluating} applied techniques for calibrating model uncertainty to each object class individually. Namely, a temperature factor or a set of binning grids (\ie, hyper-parameters) has to be estimated for each of the hundreds of classes in the LVIS dataset, leaving the techniques sensitive to hyper-parameter tuning. Indeed, \citet{dave2021evaluating} showed that it is quite challenging to estimate those hyper-parameters for tail classes. In contrast, the techniques we apply have only a single hyper-parameter, which can be selected robustly from the training data.


\section{Post-Processing Calibration for Long-Tailed Object Detection}
\label{s_approach}

In this section, we provide the background and notation for long-tailed object detection and instance segmentation, describe our approach \textbf{Normalized Calibration}
(\ourmethod), 
and discuss its relation to existing post-processing calibration methods.

\subsection{Background and Notation}
\label{ss_background}

Our tasks of interests are object detection and instance segmentation. Object detection focuses on detecting objects via bounding boxes while instance segmentation additionally requires precisely segmenting each object instance in an image. Both tasks involve classifying the object in each box/mask proposal region into one of the pre-defined classes. This classification component is what our proposed approach aims to improve. 
The most common object classification loss is the cross-entropy (CE) loss,
\begin{align}
 \mathcal{L}_{\text{CE}}(\vx, \vy) = -\sum_{c=1}^{C+1} y[c] \times\log\big( p(c|\vx)\big),\label{eq_ce}
\end{align}
where $\vy \in \{0, 1\}^{C+1}$ is the one-hot vector of the ground-truth class and $p(c|\vx)$ is the predicted probability (\ie, confidence score) of the proposal $\vx$ belonging to the class $c$, which is of the form
\begin{align}
    s_c = p(c|\vx) = 
    \frac{\exp(\phi_c(\vx))}{\sum_{c'=1}^{C}\exp(\phi_{c'}(\vx)) + \exp(\phi_{C+1}(\vx))}.\label{eq_softmax}
\end{align}
Here, $\phi_c$ is the logit for class $c$, which is usually realized by  $\vw_c^{\top}f_{\vtheta}(\vx)$: $\vw_c$ is the linear classifier associated with class $c$ and $f_{\vtheta}$ is the feature network. We use $C+1$ to denote the ``background'' class.

\begin{wrapfigure}{r}{0.45\textwidth}
    \centering
    \vspace{-3mm}
    \includegraphics[width=0.42\textwidth]{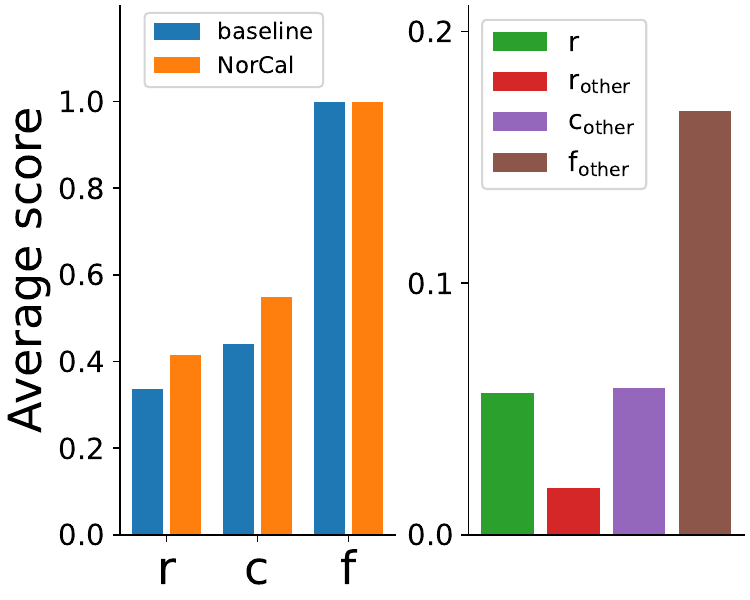}
    \vspace{-2mm}
    
    \caption{ \newadd{\small \textbf{The effect of long-tailed distributions on LVIS v0.5~\cite{gupta2019lvis}.} \textbf{Left:} We show the confidence scores of the top 300 tuples of each image by the baseline Faster R-CNN detector~\cite{ren2016faster} w/o or w/ \ourmethod. We average the scores for rare, common, and frequent classes and then linearly scale these averaged scores such that the frequent class has a score of 1. The baseline detector gives frequent objects higher scores, which can be alleviated by \ourmethod. \textbf{Right:} For tuples of the \emph{rare} class predicted by the baseline Faster R-CNN w/o \ourmethod, we further show the average scores of \textbf{\color{rare}them} and the average highest scores from another \textbf{\color{r_other}{rare}}, \textbf{\color{c_other}{common}}, and \textbf{\color{f_other}{frequent}} classes on \emph{the same proposals}. The learned baseline detector tends to predict frequent classes for a proposal.}}
    \label{fig:effect}
    \vspace{-5mm}
\end{wrapfigure}

During testing, a set of ``(box/mask proposal, object class, confidence score)'' tuples are generated for each image; each proposal can be paired with multiple classes and appears in multiple tuples if the corresponding scores are high enough.
The most common {evaluation metric} for these tuples is average precision (AP), where they are compared against the ground-truths \emph{for each class}\footnote{The difference between AP for object detection and instance segmentation lies in the computation of the intersection over union (IoU): the former based on boxes and the latter based on masks.}. Concretely, the tuples with predicted class $c$ will be gathered, sorted by their scores, and compared with the ground-truths for class $c$. 
Further, for popular benchmarks such as MSCOCO~\cite{lin2014microsoft} and LVIS~\cite{gupta2019lvis}, there is a cap $K$ (often set to $300$) on the number of detected objects per image, which is enforced usually by including only the tuples with top $K$ confidence scores.
Such a cap makes sense in practice, since a scene seldom contains over $300$ objects; creating too many, likely noisy tuples can also be annoying to users (\eg, for a camera equipped with object detection).

\mypara{Long-tailed object detection and instance segmentation: problems and empirical evidence.}
Let $N_c$ denote the number of training images of class $c$. A major challenge in long-tailed object detection is that $N_c$ is imbalanced across classes, and the learned classifier using~\autoref{eq_ce} is biased toward giving higher scores to the head classes (whose $N_c$ is larger) \cite{he2009learning,buda2018systematic,CBfocal,dave2021evaluating}. For instance, in the long-tailed object detection benchmark LVIS~\cite{gupta2019lvis} whose classes are divided into frequent ($N_c>100$), common ($100\geq N_c>10$), and rare ($N_c\leq10$), the confidence scores of the rare classes are much smaller than the frequent classes during inference (see \autoref{fig:effect}). As a result, the top $K$ tuples mostly belong to the frequent classes; proposals of the rare classes are often mis-classified as frequent classes, which aligns with the observations by \citet{wang2020devil}. 

\subsection{Normalized Calibration for Long-tailed Object Detection (\ourmethodbold)}
\label{ss_method}

\newadd{
Next, we describe the key components of the proposed \ourmethod, including confidence score calibration and normalization. The former re-ranks the confidence scores across classes to overcome the bias that rare classes usually have lower confidence scores; the latter helps to re-order the scores of detected tuples within each class for further improving the performance. Both the confidence score calibration and normalization are essential to the success of \ourmethod.
}

\mypara{Post-processing calibration and foreground-background decomposition.} 
We explore applying simple post-calibration techniques from standard multi-way classification~\cite{kang2019decoupling,wu2021adversarial} to object detection and instance segmentation. The main idea is to scale down the logit of each class $c$ by its size $N_c$~\cite{ye2020identifying,menon2020long}. 
In our case, however, the background class poses a unique challenge. First, $N_{C+1}$ is ill-defined since nearly all images contain backgrounds. Second, the background regions extracted during model training are drastically different from the foreground object proposals in terms of amounts and appearances. We thus propose to decompose \autoref{eq_softmax} as follows,
\begin{align}
    p(c|\vx) = \frac{\sum_{c'=1}^{C}\exp(\phi_{c'}(\vx))}{\sum_{c'=1}^{C}\exp(\phi_{c'}(\vx)) + \exp(\phi_{C+1}(\vx))}\times\frac{\exp(\phi_c(\vx))}{\sum_{c'=1}^{C}\exp(\phi_{c'}(\vx))},\label{eq_decompose_softmax}
\end{align}
where the first term on the right-hand side predicts how likely $\vx$ is foreground (vs. background, \ie, class $C+1$) and the second term predicts how likely $\vx$ belongs to class $c$ given that it is foreground.
Note that, the background logit $\phi_{C+1}(\vx)$ only appears in the first term and is compared to all the foreground classes as a whole. In other words, scaling or reducing it does not change the order of confidence scores among the object classes $c\in\{1,\cdots,C\}$. We thus choose to keep $\phi_{C+1}(\vx)$ intact. Please refer to \autoref{s_exp} for a detailed analysis, including 
the effect of adjusting $\phi_{C+1}(\vx)$.

For the foreground object classes, inspired by~\autoref{fig:effect} and the studies in~\cite{ye2020identifying,dave2021evaluating,menon2020long}, we propose to scale down the exponential of the logit $\phi_{c}(\vx), \forall c\in\{1,\cdots,C\}$, by a positive factor $a_c$,
\begin{align}
    & p(c|\vx) = \frac{\exp(\phi_c(\vx))/a_c}{\sum_{c'=1}^{C}\exp(\phi_{c'}(\vx))/a_{c'} + \exp(\phi_{C+1}(\vx))},
    \label{eq_post_1}
\end{align}
in which $a_c$ should monotonically increase with respect to $N_c$ --- such that the scores for head classes will be suppressed.
We investigate a simple way to set $a_c$, inspired by~\cite{ye2020identifying,kim2020adjusting},
\begin{align}
    a_c = N_c^\gamma, \hspace{5pt} \gamma \geq 0 
    \label{eq_calib_ac},
\end{align}
which has a single hyper-parameter $\gamma$ that controls the strength of dependency between $a_c$ and $N_c$. Specifically, if $\gamma=0$, we recover the original confidence scores in \autoref{eq_softmax}. We investigate other methods beyond \autoref{eq_post_1} and \autoref{eq_calib_ac} in \autoref{s_exp}.

\mypara{Hyper-parameter tuning.} Our approach only has a single hyper-parameter $\gamma$ to tune. We observe that we can tune $\gamma$ directly on the training data\footnote{Unlike imbalanced classification in which the learned classifier ultimately achieves $\sim100\%$ accuracy on the training data \cite{ye2020identifying,ye2021procrustean} (so hyper-parameter tuning using the training data becomes infeasible), a long-tailed object detector can hardly achieve $100\%$ AP per class even on the training data.}, bypassing the need of a held-out set which can be hard to collect due to the scarcity of examples for the tail classes. \citet{dave2021evaluating} also investigate this idea; however, the selected hyper-parameters from training data hurt the test results of rare classes. We attribute this to the fact that their methods have separate hyper-parameters for each class, and that makes them hard to tune.

\begin{figure}[t]
    \centering
    \includegraphics[width=1\textwidth]{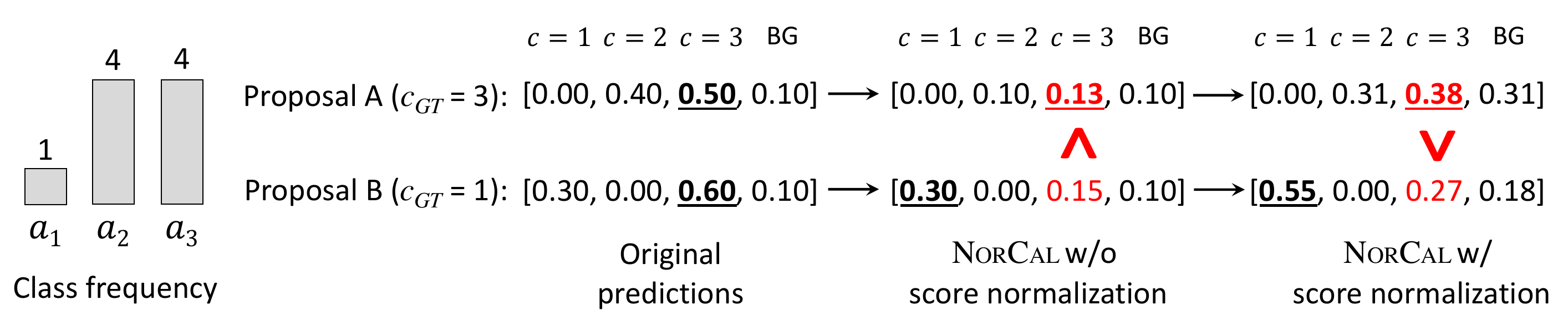}
    \vskip -3pt
    \caption{\small \textbf{\ourmethodbold with score normalization can improve AP for head classes.} Here we assume there are three possible foreground classes, and show the ground-truth classes (\ie, $c_{GT}$) and predictions for two object proposals. Bold and underlined numbers indicate the highest scored class for each proposal.
    The proposed calibration approach and score normalization can be organically coupled together to improve the ranking of personals/tuples for each class. See the text for details.}
    \label{fig:norm}
    \vskip -10pt
\end{figure}

\mypara{The importance of normalization and its effect on AP.} At first glance, our approach \ourmethod seems to simply scale down the scores for head classes, and may unavoidably hurt their AP due to the decrease of detected tuples (hence the recall) within the cap. However, we point out that the normalization operation (\ie, sum to 1) in \autoref{eq_post_1} can indeed improve AP for head classes --- \emph{normalization enables re-ordering the scores of tuples within each class.} 

Let us consider a three-class example (see \autoref{fig:norm}), in which $c=1$ is a tail class, $c=2$ and $c=3$ are head classes, and $c=4$ is the background class. Suppose two proposals are found from an image: proposal $A$ has scores $[0.0, 0.4, 0.5, 0.1]$ and the true label $c_{GT}=3$; proposal $B$ has scores $[0.3, 0.0, 0.6, 0.1]$ and the true label $c_{GT}=1$. Before calibration, proposal $B$ is ranked higher than $A$ for $c=3$, resulting in a low AP. Let us assume $a_1 = 1$ and $a_2 = a_3 = 4$. If we simply divide the scores of object classes by these factors,
proposal $B$ will still be ranked higher than $A$ for $c=3$. However, by applying \autoref{eq_post_1}, we get the new scores for proposal $A$ as $[0.0, 0.31, 0.38, 0.31]$ and for proposal $B$ as $[0.55, 0.0, 0.27, 0.18]$ --- proposal $A$ is now ranked higher than $B$ for $c=3$, leading to a higher AP for this class. As will be seen in \autoref{s_exp}, such a ``re-ranking'' property is the key to making \ourmethod excel in AP for all classes as well as in other metrics like AP$^\text{Fixed}$~\cite{dave2021evaluating}.

\subsection{Comparison to Existing Work}
\label{ss_method_analysis}

\citet{li2020overcoming} investigated classifier normalization~\cite{kang2019decoupling} for post-processing calibration. They modified the calculation of $\phi_c$ from $\vw_c^{\top}f_{\vtheta}(\vx)$ to $\frac{\vw_c^{\top}}{\|\vw_c\|_2^\gamma}f_{\vtheta}(\vx)$, building upon the observation that the classifier weights of head classes tend to exhibit larger norms~\cite{kang2019decoupling}. The results, however, were much worse than their proposed cost-sensitive method BaGS. 
They attributed the inferior result to the background class, and had combined two models, with or without classifier normalization, attempting to improve the accuracy. 
Our decomposition in \autoref{eq_decompose_softmax} suggests a more straightforward way to handle the background class. Moreover, $N_c$ provides a better signal for calibration than $\|\vw_c\|_2$, according to~\cite{ye2020identifying,kim2020adjusting}. We provide more discussions and comparison results in the {\color{magenta}{supplementary material}}.

\subsection{Extension to Multiple Binary Sigmoid Classifiers}
\label{ss_binary}
Many existing models for long-tailed object detection and instance segmentation are based on multiple binary classifiers instead of the softmax classifier~\cite{tan2020equalization,ren2020balanced,tan2020equalizationv2,wang2021adaptive,hsieh2021droploss}. That is, $s_c$ in \autoref{eq_softmax} becomes
\begin{align}
    s_c = \frac{1}{1 + \exp(-\vw_c^\top f_{\vtheta}(\vx))} = \frac{1}{1 + \exp(-\phi_c(\vx))} = \frac{\exp(\phi_c(\vx))}{\exp(\phi_c(\vx)) + 1},
    \label{eq_binary}
\end{align}
in which $\vw_c$ treats every class $c'\neq c$ and the background class together as the ``negative'' class. In other words, the background logit $\phi_{C+1} = \vw_{C+1}^\top f_{\vtheta}(\vx)$ in \autoref{eq_softmax} is not explicitly learned.

Our post-processing calibration approach can be extended to multiple binary classifiers as well. For example, \autoref{eq_post_1} becomes
\begin{align}
    s_c = \frac{\exp(\phi_c(\vx))/ a_{c}}{\exp(\phi_c(\vx))/ a_{c} + 1}.
\end{align}
\newadd{
We note that solely calibrating the scores can re-rank the detected tuples across classes within each image such that rare and common objects, which initially have lower scores, could be included in the cap to largely increase the recall. Therefore, as will be shown in the experimental results, the improvement for multiple binary classifiers mainly comes from the rare and common objects. 

However, one drawback of the score calibration alone is the infeasibility of normalization across classes; $s_c$ does not necessarily sum to 1, making it hard to re-order the scores of tuples within each class.
Forcing the confidence scores across classes of each proposal to sum to 1 would inevitably turn many background patches into foreground proposals due to the lack of the background logit $\phi_{C+1}$.}


\section{Experiments}
\label{s_exp}

\subsection{Setup}
\label{ss_exp_setup}
\mypara{Dataset.}
We validate \ourmethod on the LVIS v1 dataset~\cite{gupta2019lvis}, a benchmark dataset for large-vocabulary instance segmentation which has 100K/19.8K/19.8K training/validation/test images.  There are 1,203 categories, divided into three groups based on the number of training images per class: rare (1--10 images), common (11--100 images), and frequent ($>$100 images). 
\emph{All results are reported on the validation set \newadd{of LVIS v1}.}
\newadd{For comparisons to more existing works and different tasks, we also conduct detailed experiments and analyses on LVIS v0.5~\cite{gupta2019lvis}, Objects365~\cite{shao2019objects365}, MSCOCO~\cite{lin2014microsoft}, and image classification datasets in the {\color{magenta}{supplementary material}}.}

\mypara{Evaluation metrics.}
We adopt the standard mean \textbf{Average Precision (AP)}~\cite{lin2014microsoft} for evaluation. The cap over detected objects per image is set as $300$ (cf. \autoref{ss_background}).
Following \cite{gupta2019lvis}, we denote the mean AP for rare, common, and frequent categories by $\text{AP}_{r}$, $\text{AP}_{c}$, and $\text{AP}_{f}$, respectively. 
We also report results with a complementary metric \textbf{$\text{AP}^\text{{Fixed}}$}~\cite{dave2021evaluating}, which replaces the cap over detected objects per image by a cap over detected objects per class from the entire validation set.
Namely, $\text{AP}^\text{{Fixed}}$ removes the competition of confidence scores among classes within an image, making itself category-independent. We follow~\cite{dave2021evaluating} to set the per-class cap as $10,000$. \newadd{Instead of Mask AP, we also report the results in \textbf{$\text{AP}^\text{{Fixed}}$}~\cite{dave2021evaluating} with \textbf{Boundary IoU}, following the standard evaluation metric in LVIS Challenge 2021\footnote{\url{https://www.lvisdataset.org/challenge_2021}.}.} Meanwhile, we report \textbf{\APb}, which assesses the AP for the bounding boxes produced by the instance segmentation models.

\mypara{Implementation details and variants.} 
We apply \ourmethod to post-calibrate several representative baseline models,
for which we use the released checkpoints from the corresponding papers. We focus on models that have \emph{a softmax classifier or multiple binary classifiers} for assigning labels to proposals\footnote{Several existing methods (\eg, \cite{li2020overcoming,tan2020equalizationv2,wang2020seesaw}) develop specific classification rules to which \ourmethod cannot be directly applied.}.
For \ourmethod, (a) we investigate different mechanisms by applying post-calibration to the classifier logits, exponentials, or probabilities (cf. \autoref{eq_post_1}); 
(b) we study different types of calibration factor $a_c$, using the class-dependent temperature (CDT)~\cite{ye2020identifying} presented in \autoref{eq_calib_ac} or the effective number of samples (ENS)~\cite{CBfocal};
(c) we compare with or without score normalization. 
We tune the only hyper-parameter of \ourmethod (\ie, in $a_c$) on training data. 
 
\subsection{Main Results}
\label{ss_main_result}

\mypara{\ourmethodbold effectively improves baselines in diverse scenarios.} We first apply \ourmethod to representative baselines for instance segmentation:
(1) Mask R-CNN~\cite{he2017mask} with feature pyramid networks~\cite{lin2017feature}, which is trained with repeated factor sampling (RFS), following the standard training procedure in~\cite{gupta2019lvis}; 
(2) re-sampling/cost-sensitive based methods that have a multi-class classifier, \eg, cRT~\cite{kang2019decoupling}; 
(3) re-sampling/cost-sensitive based methods that have multiple binary classifiers, \eg, EQL~\cite{tan2020equalization};
(4) data augmentation based methods, \eg, a state-of-the-art method MosaicOS~\cite{zhang2021simple}. 
\emph{Please see the {\color{magenta}{supplementary material}} for a comparison with other existing methods.}

\begin{table*}[]
\centering
\tabcolsep 5pt
\fontsize{8}{9}\selectfont
\renewcommand\arraystretch{1.0}
\caption{\small \textbf{Comparison of instance segmentation on the validation set of LVIS v1.} \ourmethod provides solid improvement to existing models.
{\color{red}{$\dagger$}}: with EQL, we see a slight drop on the frequent classes due to the infeasibility of score normalization across classes with multiple binary classifiers. $\star$: models from \cite{zhang2021simple}. $\ddagger$: models from \cite{tan2020equalizationv2}.}
\label{tab:v1_IS}
\begin{tabular}{@{\;}l@{\quad}l@{\;}c@{\;}rrrrr@{\;}}
\toprule
{Backbone} & {Method} & {\ourmethod} & {AP} & {\AP{r}} & {\AP{c}} & {\AP{f}} & {\APb} \\
\midrule
\multirow{8}{*}{R-50~\cite{he2016deep}} & \multirow{2}{*}{EQL~\cite{tan2020equalization}$\ddagger$} & {\textcolor{codegray}{\xmark}} & 18.60 & 2.10 & 17.40 & 27.20 & {19.30} \\
& & {\textcolor{cerise}{\cmark}} & \cellcolor{LightCyan}{(+2.30) 20.90} & \cellcolor{LightCyan}{(+3.90) 6.00} & \cellcolor{LightCyan}{(+3.80) 21.20} & \cellcolor{LightCyan}{{\color{red}{$\dagger$}}(-0.10) 27.10} & \cellcolor{LightCyan}{(+2.50) 21.80}  \\
 & \multirow{2}{*}{cRT~\cite{kang2019decoupling}$\ddagger$} & {\textcolor{codegray}{\xmark}} & 22.10 & 11.90 & 20.20 & 29.00 & {22.20} \\
 & & {\textcolor{cerise}{\cmark}} & \cellcolor{LightCyan}{(+2.20) 24.30} & \cellcolor{LightCyan}{(+3.50) 15.40} & \cellcolor{LightCyan}{(+2.70) 22.90} & \cellcolor{LightCyan}{(+0.70) 29.70} & \cellcolor{LightCyan}{(+1.50) 23.70} \\
 & \multirow{2}{*}{RFS~\cite{gupta2019lvis}$\star$} & {\textcolor{codegray}{\xmark}} & 22.58 & 12.30 & 21.28 & 28.55 & 23.25 \\
 & & {\textcolor{cerise}{\cmark}} & \cellcolor{LightCyan}{(+2.65) 25.22} & \cellcolor{LightCyan}{(+7.03) 19.33} & \cellcolor{LightCyan}{(+2.88) 24.16} & \cellcolor{LightCyan}{(+0.43) 28.98} & \cellcolor{LightCyan}{(+2.83) 26.08} \\
 & \multirow{2}{*}{MosaicOS~\cite{zhang2021simple}} & {\textcolor{codegray}{\xmark}} & 24.45 & 18.17 & 22.99 & 28.83 & 25.05 \\
 & & {\textcolor{cerise}{\cmark}} & \cellcolor{LightCyan}{(+2.32) {26.76}} & \cellcolor{LightCyan}{(+5.69) {23.86}}  & \cellcolor{LightCyan}{(+2.82) {25.82}} & \cellcolor{LightCyan}{(+0.27) {29.10}} & \cellcolor{LightCyan}{(+2.73) {27.77}} \\
 \midrule
 \multirow{4}{*}{R-101~\cite{he2016deep}} 
 & \multirow{2}{*}{RFS~\cite{gupta2019lvis}$\star$} & {\textcolor{codegray}{\xmark}} & {24.82} & {15.18} & {23.71} & {30.31} & {25.45} \\
 & & {\textcolor{cerise}{\cmark}} & \cellcolor{LightCyan}{(+2.43) {27.25}} & \cellcolor{LightCyan}{(+5.61) {20.79}} & \cellcolor{LightCyan}{(+2.74) {26.45}} & \cellcolor{LightCyan}{(+0.68) {30.99}} & \cellcolor{LightCyan}{(+2.60) {28.05}} \\
 & \multirow{2}{*}{MosaicOS~\cite{zhang2021simple}} & {\textcolor{codegray}{\xmark}} & {26.73} & {20.53} & {25.78} & {30.53} & {27.41} \\
 & & {\textcolor{cerise}{\cmark}} & \cellcolor{LightCyan}{(+2.30) {29.03}} & \cellcolor{LightCyan}{(+5.85) {26.38}} & \cellcolor{LightCyan}{(+2.37) {28.15}} & \cellcolor{LightCyan}{(+0.66) {31.19}} & \cellcolor{LightCyan}{(+2.55) {29.96}} \\
 \midrule
 \multirow{4}{*}{X-101~\cite{xie2017aggregated}}  & \multirow{2}{*}{RFS~\cite{gupta2019lvis}$\star$} & {\textcolor{codegray}{\xmark}} & {26.67} & {17.60} & {25.58} & {31.89} & {27.35} \\
 & & {\textcolor{cerise}{\cmark}} & \cellcolor{LightCyan}{(+1.25) {27.92}} & \cellcolor{LightCyan}{(+2.15) {19.75}} & \cellcolor{LightCyan}{(+1.61) {27.19}} & \cellcolor{LightCyan}{(+0.45) {32.34}} & \cellcolor{LightCyan}{(+1.49) {28.83}} \\
 & \multirow{2}{*}{MosaicOS~\cite{zhang2021simple}} & {\textcolor{codegray}{\xmark}} & {28.29} & {21.75} & {27.22} & {32.35} & {28.85} \\
 & & {\textcolor{cerise}{\cmark}} & \cellcolor{LightCyan}{(+1.52) {29.81}} & \cellcolor{LightCyan}{(+3.97) {25.72}} & \cellcolor{LightCyan}{(+1.70) {28.92}} & \cellcolor{LightCyan}{(+0.24) {32.59}} & \cellcolor{LightCyan}{(+1.71) {30.56}} \\
 \bottomrule
 \vspace{-5mm}
\end{tabular}
\end{table*}


\autoref{tab:v1_IS} provides our main results on LVIS v1. \ourmethod achieves consistent gains on top of all the models of different backbone architectures. 
For instance, for RFS~\cite{gupta2019lvis} with ResNet-50, the overall AP improves from 22.58$\%$ to 25.22$\%$, including $\sim7\%/3\%$ gains on $\text{AP}_{r}/\text{AP}_{c}$ for rare/common objects.
Importantly, we note that \ourmethod's improvement is on almost all the evaluation metrics (columns), demonstrating a key strength of \ourmethod that is not commonly seen in literature: achieving overall gains without sacrificing the $\text{AP}_f$ on frequent classes. We attribute this to the score normalization operation of \ourmethod: unlike~\cite{dave2021evaluating} which only re-ranks scores across categories, \ourmethod further re-ranks the scores within each category. 
Indeed, the only performance drop in \autoref{tab:v1_IS} is on frequent classes for EQL, which is equipped with multiple binary classifiers such that score normalization across classes is infeasible (cf. \autoref{ss_binary}). 
We provide more discussions in the ablation studies.

\mypara{Comparison to existing post-calibration methods.}
We then compare our \ourmethod to other post-calibration techniques. Specifically, we compare to those in~\cite{dave2021evaluating} on the LVIS v1 instance segmentation task, including Histogram Binning~\cite{zadrozny2001obtaining}, Bayesian binning into quantiles (BBQ)~\cite{naeini2015obtaining}, Beta calibration~\cite{kull2017beta}, isotonic regression~\cite{zadrozny2002transforming}, and Platt scaling~\cite{platt1999probabilistic}. We also compare to classifier normalization ($\tau$-normalized)~\cite{li2020overcoming,kang2019decoupling} on the LVIS v0.5 object detection task. All the hyper-parameters for calibration are tuned from the training data.

\autoref{tab:comp_exist_post_calib} shows the results. \ourmethod significantly outperforms other techniques on both tasks and can improve AP for all the classes. We attribute the improvement over methods studied in \cite{dave2021evaluating} to two reasons: first, \ourmethod has only one hyper-parameter, while calibration methods in \cite{dave2021evaluating} have hyper-parameters for every category and thus are sensitive to tune; second, \ourmethod performs score normalization, while \cite{dave2021evaluating} does not. Compared to \cite{li2020overcoming,kang2019decoupling}, the use of per-class data count in \ourmethod has been shown to outperform classifier norms for calibrating classifiers~\cite{ye2020identifying,kim2020adjusting}.

\begin{table}
	\begin{minipage}[t]{.43\textwidth}
	    \centering
        \tabcolsep 2.6pt
        \fontsize{8}{9}\selectfont
        \renewcommand\arraystretch{1.32}
        \caption{\small \textbf{Comparison to other existing post-calibration methods.} \ourmethod outperforms methods studied in \cite{dave2021evaluating} and \cite{li2020overcoming}. $\dagger$: w/o RFS~\cite{gupta2019lvis}.}
        \label{tab:comp_exist_post_calib}
        \begin{tabular}{lcccc}
        \toprule
        \multicolumn{1}{l}{Segmentation on v1} & $\text{AP}$ & $\text{AP}_{r}$  & {$\text{AP}_{c}$} & {$\text{AP}_{f}$} \\
        \midrule
        RFS~\cite{gupta2019lvis} & 22.58 & 12.30 & 21.28 & 28.55 \\
        w/ HistBin~\cite{zadrozny2001obtaining} & 21.82 & 11.28 & 20.31 & 28.13 \\
        w/ BBQ (AIC)~\cite{naeini2015obtaining} & 22.05 & 11.41 & 20.72 & 28.21 \\
        w/ Beta calibration~\cite{kull2017beta} & 22.55 & 12.29 & 21.27 & 28.49 \\
        w/ Isotonic reg.~\cite{zadrozny2002transforming} & 22.43 & 12.19 & 21.12 & 28.41 \\
        w/ Platt scaling~\cite{platt1999probabilistic} & 22.55 & 12.29 & 21.27 & 28.49 \\
        \cellcolor{LightCyan}{w/ \ourmethod} & \cellcolor{LightCyan}{25.22} & \cellcolor{LightCyan}{19.33} & \cellcolor{LightCyan}{24.16} & \cellcolor{LightCyan}{28.98} \\
        \bottomrule
        \toprule
        \multicolumn{1}{l}{Detection on v0.5} & $\text{AP}^{b}$ & $\text{AP}^{b}_{r}$ & $\text{AP}^{b}_{c}$ & $\text{AP}^{b}_{f}$ \\
        \midrule
        Faster R-CNN~\cite{ren2016faster}$\dagger$ & 20.98 & 4.13 & 19.70 & 29.30 \\
        w/ $\tau$-normalized~\cite{li2020overcoming}$\dagger$ & 21.61 & 6.18 & 20.99 & 28.54 \\
        \cellcolor{LightCyan}{w/ \ourmethod$\dagger$} &\cellcolor{LightCyan}{23.87}	&\cellcolor{LightCyan}{6.98}	&\cellcolor{LightCyan}{24.17}	&\cellcolor{LightCyan}{30.24} \\
        \bottomrule
    \end{tabular}
        
	\end{minipage}
	\hfill
	\begin{minipage}[t]{.53\textwidth}
	
	\centering
        \fontsize{8}{9}\selectfont
        \tabcolsep 2.5pt
        \renewcommand\arraystretch{1.0}
        \centering
        \caption{\small 
            \textbf{Ablation studies of \ourmethod with various modeling choices and mechanisms.} We report results on LVIS v1 instance segmentation. \cali: calibration mechanism. \nor: class score normalization. The best ones are in bold.
        }
        \label{tab:ablation}
        \begin{tabular}{ccccccc}
        \toprule
        {$a_c$} & {\cali} & {\nor} & {AP} & {\AP{r}} & {\AP{c}} & {\AP{f}} \\
        \midrule
        \multicolumn{1}{c}{Baseline} &  $\exp(\phi_c(\vx))$ &  {\textcolor{cerise}{\cmark}} & 22.58 & 12.30 & 21.28 & 28.55 \\
        \midrule
        \multirow{7}{*}{\makecell[c]{$\cfrac{1 - \gamma^{N_c}}{1 - \gamma}$ \\ (ENS~\cite{CBfocal})}} & \multirow{2}{*}{$\exp(\phi_c(\vx)/a_c)$} & {\textcolor{codegray}{\xmark}} & 23.66 & 14.55 & 22.36 & \textbf{29.11}  \\
         &  & {\textcolor{cerise}{\cmark}}  & 23.96 & 15.84 & 22.61 & 29.04 \\
        \cmidrule(r){2-7}
         & \multirow{2}{*}{$p(c|\vx)/a_c$} & {{\textcolor{codegray}{\xmark}}} & 24.18 & 18.88 & 22.95 & 27.87 \\
         &  & {\textcolor{cerise}{\cmark}} & 24.85 & \textbf{19.43} & 23.67 & 28.54 \\
         \cmidrule(r){2-7}
         & \multirow{2}{*}{$\exp(\phi_c(\vx))/a_c$} & {{\textcolor{codegray}{\xmark}}} & 17.49 & 14.16 & 17.20 & 19.27 \\
         &  & {\textcolor{cerise}{\cmark}} & 24.85 & \textbf{19.43} & 23.67 & 28.54 \\
        \midrule
        \multirow{7}{*}{\makecell[c]{$N_{c}^{\gamma}$\\(CDT~\cite{ye2020identifying})}} & \multirow{2}{*}{$\exp(\phi_c(\vx)/a_c)$} & {{\textcolor{codegray}{\xmark}}} & 17.52 & 14.04 & 17.38 & 19.20 \\
         &  & {\textcolor{cerise}{\cmark}} & 24.77 & 17.99 & 23.81 & 28.83 \\
        \cmidrule(r){2-7}
         & \multirow{2}{*}{$p(c|\vx)/a_c$} & {{\textcolor{codegray}{\xmark}}} & 24.50 & 18.34 & 23.42 & 28.41 \\
         &  & {\textcolor{cerise}{\cmark}} & \textbf{25.22} & 19.33 & \textbf{24.16} & 28.98 \\
         \cmidrule(r){2-7}
         & \multirow{2}{*}{$\exp(\phi_c(\vx))/a_c$} & {{\textcolor{codegray}{\xmark}}} & 17.52 & 13.93 & 17.24 & 19.41 \\
         &  & {\textcolor{cerise}{\cmark}} & \cellcolor{LightCyan}{\textbf{25.22}} & \cellcolor{LightCyan}{19.33} & \cellcolor{LightCyan}{\textbf{24.16}} & \cellcolor{LightCyan}{28.98} \\
         \bottomrule
        \end{tabular}

	\end{minipage}
\end{table}


\subsection{Ablation Studies and Analysis}
\label{ss_as_a}

We mainly conduct the ablation studies on the Mask R-CNN model~\cite{he2017mask} (with ResNet-50 backbone~\cite{he2016deep} and feature pyramid networks~\cite{lin2017feature}), trained with repeated factor sampling (RFS)~\cite{gupta2019lvis}.

\mypara{Effect of calibration mechanisms.}
In addition to reducing the logits, \ie, scaling down their exponentials (\ie, $\exp(\phi_c(\vx))/a_c$ in \autoref{eq_post_1}), we investigate another two ways of score calibration. Specifically, we scale down the output logits from the network (\ie, $\phi_c(\vx)/a_c$) or the probabilities from the classifier (\ie, $p(c|\vx)/a_c$). Again, we keep the background class intact and apply score normalization. In \autoref{tab:ablation}, we see that scaling down the exponentials and probabilities perform the same\footnote{With class score normalization, they are mathematically the same.} and outperform scaling down logits. We note that, logits can be negative; thus, scaling them down might instead increases the scores. In contrast, exponentials and probabilities are non-negative, scaling them down thus are guaranteed to reduce the scores of frequent classes more than rare classes.

\mypara{Effect of calibration factors $a_c$.}
Beyond the class-dependent temperature (CDT)~\cite{ye2020identifying} presented in \autoref{eq_calib_ac}, we study an alternative factor, inspired by the effective number of samples (ENS)~\cite{CBfocal}. Specifically, we study $a_c=(1 - \gamma^{N_c})/(1 - \gamma)$ with $\gamma\in [0, 1)$. Same as CDT, ENS has a single hyper-parameter $\gamma$ that controls the degree of dependency between $a_c$ and $N_c$. If $\gamma=0$, we recover the original confidence scores. We report the comparison of these two calibration factors in \autoref{tab:ablation}. With appropriate post-calibration mechanisms, both provide consistent gains over the baseline model.

\mypara{Importance of score normalization.}
Again in \autoref{tab:ablation}, we compare \ourmethod with or without score normalization across classes. That is, whether we include the denominator in \autoref{eq_post_1} or not.
By applying normalization, we see that \ourmethod can improve all categories, including frequent objects. 
Moreover, it is applicable to different types of calibration mechanisms as well as calibration factors.
In contrast, the results without normalization degrade at frequent classes and sometimes even at common and rare classes. We attribute this to two reasons: first, score normalization enables the detected tuples of each class to be re-ranked (cf. \autoref{fig:norm}); second, with the background logits in the denominator, the calibrated and normalized scores can effectively prevent background patches from being classified into foreground objects. 
Please be referred to the {\color{magenta}{supplementary material}} for additional results and ablation studies on sigmoid-based detectors (\ie, BALMS~\cite{ren2020balanced} and RetinaNet~\cite{lin2017focal}).

\begin{figure}[t]
	\begin{minipage}[t]{.6\textwidth}
	    \centering
        {\includegraphics[width=1\linewidth]{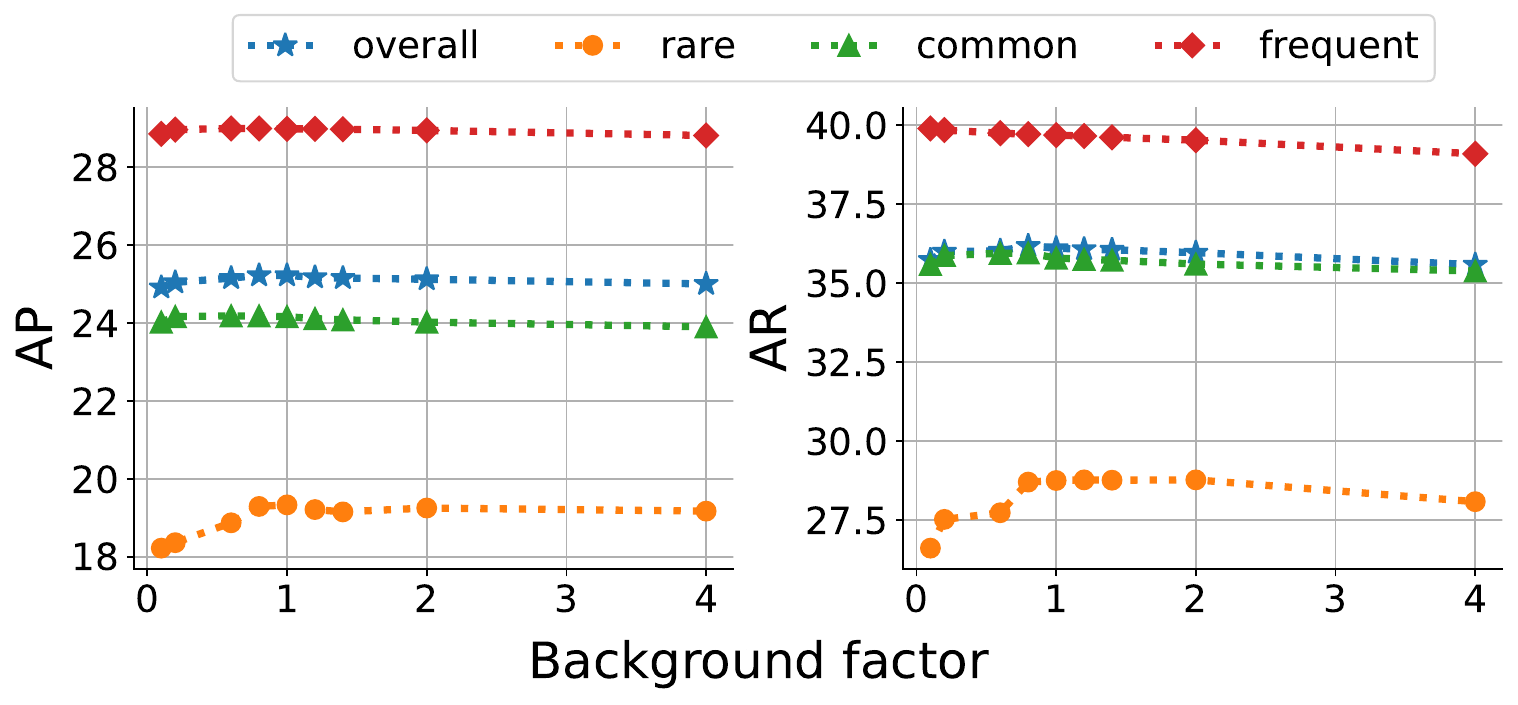}}
        \vspace{-5mm}
        \caption{\small \textbf{Results of precision and recall by adjusting background class scores.} Results are on v1 instance segmentation.}
        \label{fig:background}
	\end{minipage}
	\hfill
	\begin{minipage}[t]{.36\textwidth}
	    \centering
        {\includegraphics[width=1\textwidth]{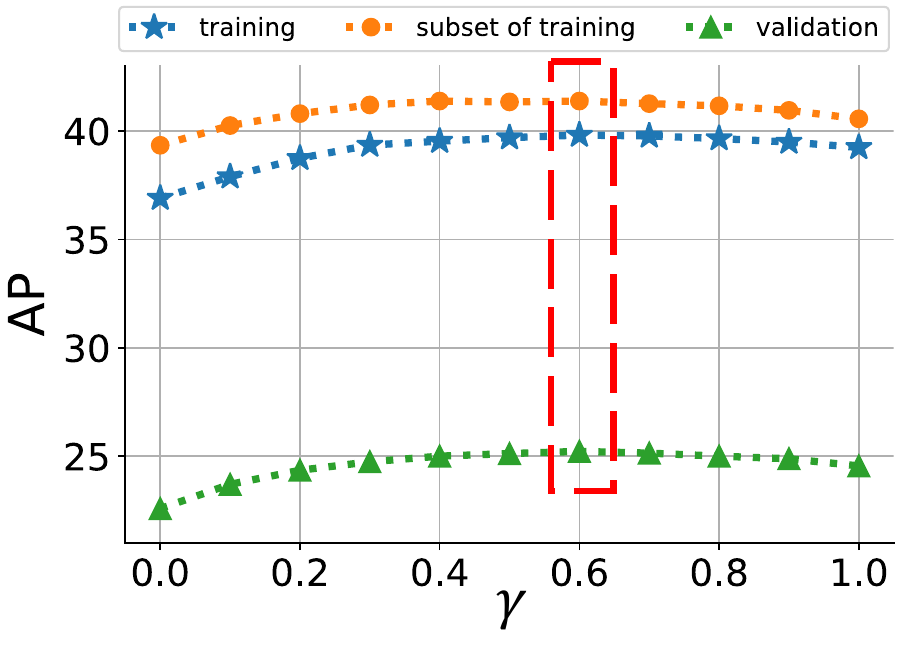}}
        \vspace{-5mm}
        \caption{\small \textbf{Calibration factor $\gamma$ can be robustly tuned using training data.}}
        \label{fig:hyper}
	\end{minipage}
	\vspace{-3mm}
\end{figure}

\begin{table}
\tabcolsep 7.8pt
\centering
\fontsize{8}{9}\selectfont
    \caption{\small \textbf{\ourmethodbold improves average precision and recall.} Results are on LVIS v1 instance segmentation.}    
    \label{tab:recall}
     \begin{tabular}{lcccc}
         \multicolumn{5}{c}{{(a) Average Precision (AP)}}\\
        \toprule
         & $\text{AP}$ &$\text{AP}_r$ & $\text{AP}_c$ & $\text{AP}_f$ \\
        \midrule
        RFS~\cite{gupta2019lvis} & 22.58 & 12.30 & 21.28 & 28.55 \\
        w/ \ourmethod & 25.22 & 19.33 & 24.16 & 28.98 \\
        \bottomrule
    \end{tabular}
    \hfill
    \begin{tabular}{lcccc}
        \multicolumn{5}{c}{{(b) Average Recall (AR)}}\\
        \toprule
        & $\text{AR}$ &$\text{AR}_r$ & $\text{AR}_c$ & $\text{AR}_f$ \\
        \midrule
        RFS~\cite{gupta2019lvis} & 30.61 & 13.73 & 28.64 & 40.24 \\
        w/ \ourmethod & 36.10 & 28.75 & 35.79 & 39.68 \\
        \bottomrule
    \end{tabular}
    \vspace{-3mm}
\end{table}

\begin{table}[t!]
\tabcolsep 7.8pt
\centering
\fontsize{8}{9}\selectfont
\caption{\small \newadd{\textbf{\ourmethodbold can improve the baseline model in $\text{AP}^\text{Fixed}$ and $\text{AP}^\text{Fixed}$ with Boundary IoU.} The baseline model uses ResNet-50 as the backbone with RFS~\cite{gupta2019lvis}. Results are reported on LVIS v1 instance segmentation.}}
    \label{tab:ap_fixed}
     \begin{tabular}{lcccc}
         \multicolumn{5}{c}{{(a) AP Fixed}}\\
        \toprule
         & $\text{AP}$ &$\text{AP}_r$ & $\text{AP}_c$ & $\text{AP}_f$ \\
        \midrule
        RFS~\cite{gupta2019lvis}  & 25.68 & 20.07 & 24.82 & 29.11 \\
        w/ \ourmethod & 26.26 & 20.56 & 25.39 & 29.73 \\
        \bottomrule
    \end{tabular}
    \hfill
    \begin{tabular}{lcccc}
        \multicolumn{5}{c}{{(b) AP Fixed with Boundary IoU}}\\
        \toprule
        & $\text{AP}$ &$\text{AP}_r$ & $\text{AP}_c$ & $\text{AP}_f$ \\
        \midrule
        RFS~\cite{gupta2019lvis} & 19.88 & 14.76 & 19.32 & 22.76 \\
        w/ \ourmethod & 20.25 & 14.99 & 19.77 & 23.10 \\
        \bottomrule
    \end{tabular}
    \vspace{-3mm}
\end{table}

\mypara{How to handle the background class?} 
\ourmethod does not calibrate the background class logit. We ablate this design by \emph{multiplying} the exponential of the background logit with a background calibration factor $\beta$, \ie, $\exp(\phi_{C+1}(\vx))\times\beta$. If $\beta=1$, there is no calibration on background class.
\autoref{fig:background} shows the average precision and recall of the model with \ourmethod w.r.t different $\beta$. 
We see consistent performance for $\beta\geq1$. For $\beta<1$, the average precision drops along with reduced $\beta$, especially for the rare classes whose average recall also drops. We note that, in the extreme case with $\beta=0$, the background class will not contribute to the final calibrated score. Thus, many background patches may be classified as foregrounds and ranked higher than rare proposals. These results and explanation justifies one key ingredient of \ourmethod  --- keeping the background logit intact.

\mypara{Sensitivity to the calibration factor.}
\ourmethod has one hyper-parameter: $\gamma$ in the calibration factor $a_c$, which controls the strength of calibration. We find that this can be tuned robustly on the \emph{training} data, even on a 5K subset of \emph{training} images: as shown in \autoref{fig:hyper}, the AP trends on the training and validation sets at different $\gamma$ are close to each other. In our experiments, we find that this observation applies to different models and backbone architectures.


\mypara{\ourmethodbold reduces false positives and re-ranks predictions within each class.}
In \autoref{tab:recall}, we show that \ourmethod can improve the AR for all classes but frequent objects (with a slight drop). The gains on AP for frequent classes thus suggest that \ourmethod can re-rank the detected tuples within each class, pushing many false positives to have scores lower than true positives.


\newadd{
\mypara{\ourmethodbold is effective in $\text{AP}^\text{Fixed}$~\cite{dave2021evaluating} and Boundary IoU~\cite{cheng2021boundary}.}
\autoref{tab:ap_fixed} reports the results in $\text{AP}^\text{Fixed}$ and $\text{AP}^\text{Fixed}$ with Boundary IoU. We see that \ourmethod is metric-agnostic and can consistently improve the baseline model in all groups of categories. It suggests that the improvements are due to both across-class and within-class re-ranking.  

}

\begin{figure}[t]
    \includegraphics[width=1\textwidth]{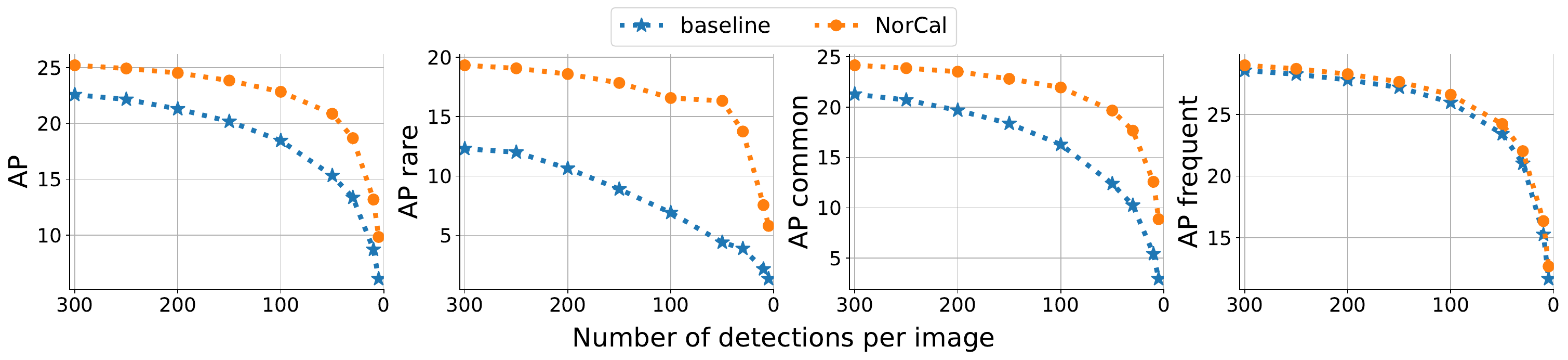}
    \caption{\small \textbf{Limits on the number of detections per image.} To perform well in a small cap, a model must rank true positives higher such that they can be included in the cap. \ourmethod performs much better than the baseline.
    }
    \label{fig:det_img2}
\end{figure}

\mypara{Limiting detections per image.}
Finally, we evaluate \ourmethod by changing the cap on the number of detections per image. 
Specifically, we investigate reducing the default number of 300. The rationale is that an image seldom contains over 300 objects. Indeed, each  LVIS~\cite{gupta2019lvis} image is annotated with around 12 object instances on average. We note that, to perform well in a smaller cap requires a model to rank most true positives in the front such that they can be included in the cap. 
In \autoref{fig:det_img2}, \ourmethod shows superior performance against the baseline model under all settings. It is worth noting that \ourmethod achieves better performance even using a strict 100 detections per image than the baseline model with 300.

\begin{figure}[t]
    \centering
    \includegraphics[width=1\textwidth]{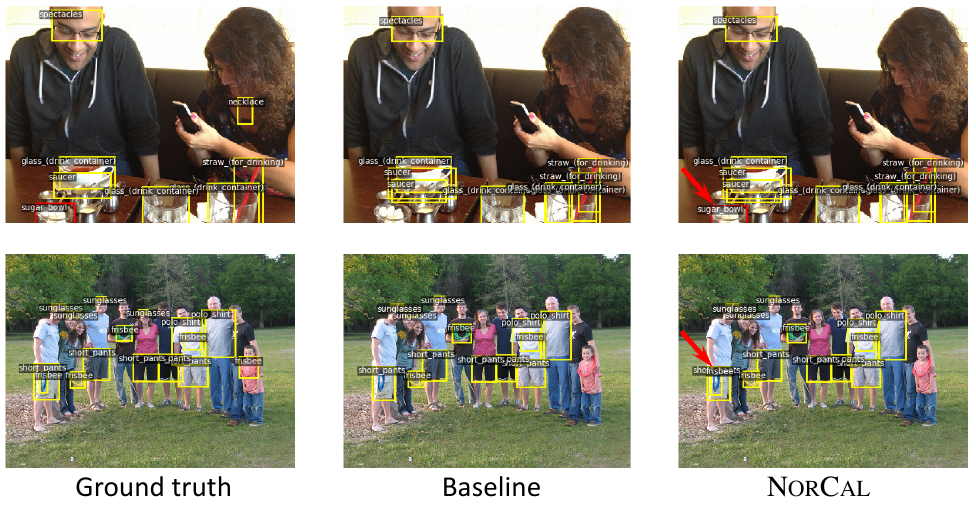}
    \caption{\small \textbf{Qualitative results.} We superimpose {\color{red}{red}} arrows to show the improvement, and  
    {\color{yellow}{Yellow}} and {\color{red}{red}} boxes to indicate the ground truth labels of frequent and rare classes. In the first example, \ourmethod successfully detects the rare object \emph{sugar bowl} without sacrificing other predictions. In the second example, even surprisingly, it can detect a missed frequent object \emph{frisbee} by the baseline. }
    \label{fig:qual}
\end{figure}

\mypara{Qualitative results.}
We show qualitative bounding box results on LVIS v1 in \autoref{fig:qual}. We compare the ground truths, the results of the baseline, and the results of \ourmethod. \ourmethod can not only detect more objects from the rare categories that may be overlooked by the baseline detector, but also improve the detection results on frequent objects. For instance, in the upper example of \autoref{fig:qual}, \ourmethod discovers a rare object ``sugar bowl'' without sacrificing any other frequent objects. Moreover, \ourmethod can improve the frequent classes, as shown in the bottom example of \autoref{fig:qual}. Please see the {\color{magenta}{supplementary material}} for more qualitative results.

\section{Conclusion}
\label{s_disc}
We present a post-processing calibration method called \ourmethod for addressing long-tailed object detection and instance segmentation. Our method is simple yet effective, requires no re-training of the already trained models, and can be compatible with many existing models to further boost the state of the art.
We conduct extensive experiments to demonstrate the effectiveness of our method in diverse settings, as well as to validate our design choices and analyze our method's mechanisms. We hope that our results and insights can encourage more future works on exploring the power of post-processing calibration in long-tailed object detection and instance segmentation.

\newadd{
\section*{Acknowledgments and Funding Transparency Statement}
This research is partially supported by NSF IIS-2107077 and the OSU GI Development funds.
We are thankful for the generous support of the computational resources by the Ohio Supercomputer Center.
We thank Zhiyun Lu  (Google) for feedback on an early draft of this paper and Han-Jia Ye (Nanjing University) for the help on image classification experiments.
}

{\small
\bibliographystyle{plainnat}
\bibliography{main}
}

    \clearpage
    \appendix
    \begin{center}
    \textbf{\Large Supplementary Material}
    \end{center}
    In this supplementary material, we provide details and additional results omitted in the main texts. 
    \begin{itemize}[leftmargin=6mm]
        \item[$\bullet$] \autoref{supp:s_related}: additional discussion on related work (\autoref{s_related} of the main paper).
        \item[$\bullet$] \autoref{suppl:s_impl}: details of experimental setups (\autoref{ss_exp_setup} of the main paper).
        \item[$\bullet$] \autoref{supp:s_results}: additional results and analysis (\autoref{ss_main_result} and \autoref{ss_as_a} of the main paper).
        \begin{itemize}
            \item[$\circ$] \autoref{supp:ss_v1_seg}: results on LVIS v1 instance segmentation.
            \item[$\circ$] \autoref{supp:ss_v0.5_seg}: results on LVIS v0.5 instance segmentation.
            \item[$\circ$] \autoref{supp:ss_v0.5_det}: results on LVIS v0.5 object detection.
            \item[$\circ$] \autoref{supp:ss_object365}: \newadd{results on Objects365 dataset.}
            \item[$\circ$] \autoref{supp:ss_mscoco}: results on MSCOCO dataset.
            \item[$\circ$] \autoref{supp:ss_classification}: \newadd{results on image classification tasks.}
            \item[$\circ$] \autoref{supp:ss_ablation}: ablation studies on sigmoid-based detectors.
            \item[$\circ$] \autoref{supp:ss_weight_norm}: further comparisons between $N_c$ and $\|\vw_c\|_2$ for \ourmethod.
            \item[$\circ$] \autoref{supp:ss_post_processing}: further analysis on existing post-processing calibration methods.
            \item[$\circ$] \autoref{supp:ss_qual}: additional qualitative results.
        \end{itemize}
    \end{itemize}
    
\section{Additional Discussion on Related Work}
\label{supp:s_related}
\subsection{Long-Tailed Object Detection and Instance Segmentation}
    Existing works can be categorized into {re-sampling}, {cost-sensitive learning}, and {data augmentation}.
    
    \mypara{Re-sampling} changes the training data distribution --- by sampling rare class data more often than frequent class ones --- to mitigate the long-tailed distribution. Re-sampling is widely adopted as a simple but effective baseline approach~\cite{gupta2019lvis,chang2021image,shen2016relay}. For example, repeat factor sampling (RFS) \cite{gupta2019lvis} sets a repeat factor (\ie, sampling frequency) for each image based on the rarest object within that image; class-aware sampling~\cite{shen2016relay} samples a uniform amount of images per class for each mini-batch. Since an image can contain multiple object classes, \citet{chang2021image} proposed to re-sample on both the image and object instance levels. RFS is the baseline approach used for the LVIS dataset~\cite{gupta2019lvis}.
    
    \mypara{Cost-sensitive learning} is the most popular category, which adjusts the cost of mis-classifying an instance or the loss of learning from an instance according to its true class label. Re-weighting is the simplest method of this kind, which gives each instance a class-specific weight in calculating the total loss (usually, tail classes with larger weights). 
    The equalization loss (EQL)~\cite{tan2020equalization} and EQL v2~\cite{tan2020equalizationv2} ignore the negative gradients for rare class classifiers or equalize the positive-negative gradient ratio for each class to balance the training, respectively. The drop loss~\cite{hsieh2021droploss} improves EQL by specifically handling the background class via re-weighting. The seesaw loss~\cite{wang2020seesaw} proposes a re-weighting scheme by combining the dataset statistics and training dynamics. Forest R-CNN \cite{wu2020forest} leverages the class hierarchical for knowledge transfer and introduces new losses for hierarchical classification.
    
    Instead of applying the new loss functions during the entire training phase, several recent methods decouple the training phase into two stages~\cite{wang2020frustratingly,kang2019decoupling,wang2020devil,hu2020learning,ren2020balanced,li2020overcoming,zhang2021distribution,wang2021adaptive}. 
    At the first stage, the object detector is trained normally just like on a relatively balanced dataset such as MSCOCO~\cite{lin2014microsoft}. Then in the second stage, re-sampling or cost-sensitive learning is employed, usually for re-training or fine-tuning only the classification network.
    Such a pipeline is shown to learn both better features and classifier. For example, two-stage fine-tuning approach (TFA)~\cite{wang2020frustratingly} first trains a base detector using only common and frequent classes, and then fine-tune the classifier and box regressor with re-sampling. Similar ideas are adopted in classifier re-training (cRT)~\cite{kang2019decoupling}, SimCal~\cite{wang2020devil}, balanced softmax (BSM)~\cite{ren2020balanced}, balanced group softmax (BaGS)~\cite{li2020overcoming}, DisAlign~\cite{zhang2021distribution}, and ACSL~\cite{wang2021adaptive}, which develop strategies or losses to re-train the classifier. Learning to segment the tail (LST)~\cite{hu2020learning} takes an incremental learning approach to gradually learn from the head to tail classes in multiple stages.
    
    \mypara{Data augmentation} improves long-tailed object detection by augmenting data for the tail classes.
    DLWL~\cite{ramanathan2020dlwl} and MosaicOS~\cite{zhang2021simple} leveraged weakly-supervised data from YFCC-100M~\cite{thomee2015yfcc100m}, ImageNet~\cite{deng2009imagenet}, and  Internet to augment the long-tailed LVIS dataset~\cite{gupta2019lvis}. 
    Copy-Paste~\cite{ghiasi2020simple} self-augments the LVIS dataset by copying object instances from one image and paste to the others.
    Instead of augmenting images, FASA~\cite{zang2021fasa} generates class-wise virtual features using a Gaussian prior whose parameters are estimated from features of real data.
    
    \subsection{Calibration of Model Uncertainty}
    \label{ss_calibration_uncertainty}
    We note that, the calibration rules we apply are different from the ones used for calibrating model uncertainty~\cite{guo2017calibration}: we aim to adjust the prediction across classes, while the latter adjusts the predicted probability to reflect the true correctness likelihood. For calibrating model uncertainty, representative methods are Platt scaling~\cite{platt1999probabilistic}, histogram binning~\cite{zadrozny2001obtaining}, Bayesian binning into quantiles (BBQ)~\cite{naeini2015obtaining}, isotonic regression~\cite{zadrozny2002transforming}, temperature scaling~\cite{guo2017calibration}, beta and Dirichlet calibration~\cite{kull2017beta,kull2019beyond}, etc.
\section{Experimental Setups}
\label{suppl:s_impl}

\subsection{Baseline Methods}
\label{suppl:ss_impl_baseline}
Our approach \ourmethod is model-agnostic as long as the detector has \textbf{\emph{a softmax classifier or multiple binary sigmoid classifiers for the objects and the background}}. Thus, we focus on those methods as long as the pre-trained models are applicable and public:
\begin{itemize}[leftmargin=5mm]
    \item The baseline Mask R-CNN~\cite{he2017mask} model with feature pyramid networks~\cite{lin2017feature}, which is trained with repeated factor sampling (RFS), following the standard training procedure in~\cite{gupta2019lvis}.
    
    \item Re-sampling/cost-sensitive based methods that have a multi-class classifier for the foreground objects and the background class, \eg, cRT~\cite{kang2019decoupling} and TFA~\cite{wang2020frustratingly}.
    
    \item Re-sampling/cost-sensitive based methods that have multiple binary sigmoid-based classifiers, \eg, EQL~\cite{tan2020equalization}, BALMS~\cite{ren2020balanced}, and \newadd{RetinaNet with focal loss~\cite{lin2017focal}}.
    
    \item Data augmentation based methods, \eg, MosaicOS~\cite{zhang2021simple}. MosaicOS augments LVIS with images from ImageNet~\cite{deng2009imagenet}, which can improve the feature network of an object detector like Faster R-CNN~\cite{ren2016faster} or Mask R-CNN~\cite{he2009learning}.
\end{itemize}

We note that, several methods change the decision/classification rules. For example, EQL v2~\cite{tan2020equalizationv2} and Seesaw~\cite{wang2020seesaw} adopt a separate background or objectness branch during the training and inference. Some other methods (BaGS~\cite{li2020overcoming} and Forest R-CNN~\cite{wu2020forest}) re-organize the category groups and apply either a group-based softmax classifier or hierarchical classification. Therefore, it is not immediately obvious how to apply calibration to them. 

\begin{table}[t]
\centering
\tabcolsep 2pt
\fontsize{8}{9}\selectfont
\renewcommand\arraystretch{1.0}
\caption{\small \textbf{Instance segmentation results on the validation set of LVIS v1.} Our method \ourmethod can improve all baseline models with different backbones to which it is applied. Seesaw~\cite{wang2020seesaw} applies a stronger ${2\times}$ training schedule while other methods are with ${1\times}$ schedule. {\color{red}{$\dagger$}}: slight performance drop on sigmoid-based detectors. $\star$: models from \cite{zhang2021simple}. $\ddagger$: models from \cite{tan2020equalizationv2}. $\clubsuit$: results from \cite{ghiasi2020simple}.}
\label{tab:suppl_v1_seg}
\begin{tabular}{clcrrrrr}
\toprule
{Backbone} & {Method} & {\ourmethod} & {AP} & {\AP{r}} & {\AP{c}} & {\AP{f}} & {\APb} \\
\midrule
\multirow{19}{*}{R-50} & DropLoss~\cite{hsieh2021droploss} &  & 19.80 & 3.50 & 20.00 & 26.70 & 20.40 \\
 & BaGS~\cite{li2020overcoming} &  & 23.10 & 13.10 & 22.50 & 28.20 & 25.76 \\
 & Forest R-CNN~\cite{wu2020forest} &  & 23.20 & 14.20 & 22.70 & 27.70 & 24.60 \\
 & RIO~\cite{chang2021image} &  & 23.70 & 15.20 & 22.50 & 28.80 & 24.10 \\
 & EQL v2~\cite{tan2020equalizationv2} &  & 23.70 & 14.90 & 22.80 & 28.60 & 24.20 \\
 & DisAlign~\cite{zhang2021distribution} &  & 24.30 & 8.50 & \textbf{26.30} & 28.10 & 23.90 \\
 & Seesaw~\cite{wang2020seesaw}$^{2\times}$ &  & 25.40 & 15.90 & 24.70 & \textbf{30.40} & 25.60 \\
 & Seesaw w/ RFS~\cite{wang2020seesaw}$^{2\times}$ &  & 26.40 & 19.60 & 26.10 & 29.80 & 27.40 \\
 \cmidrule(r){2-8}
 & \multirow{2}{*}{EQL~\cite{tan2020equalization}$\ddagger$} &  & 18.60 & 2.10 & 17.40 & 27.20 & {19.30} \\
& & {\textcolor{cerise}{\cmark}} & \cellcolor{LightCyan}{(+2.30) 20.90} & \cellcolor{LightCyan}{(+3.90) 6.00} & \cellcolor{LightCyan}{(+3.80) 21.20} & \cellcolor{LightCyan}{{\color{red}{$\dagger$}}(-0.10) 27.10} & \cellcolor{LightCyan}{(+2.50) 21.80}  \\
 & \multirow{2}{*}{cRT~\cite{kang2019decoupling}$\ddagger$} &  & 22.10 & 11.90 & 20.20 & 29.00 & {22.20} \\
 & & {\textcolor{cerise}{\cmark}} & \cellcolor{LightCyan}{(+2.20) 24.30} & \cellcolor{LightCyan}{(+3.50) 15.40} & \cellcolor{LightCyan}{(+2.70) 22.90} & \cellcolor{LightCyan}{(+0.70) 29.70} & \cellcolor{LightCyan}{(+1.50) 23.70} \\
 & \multirow{2}{*}{RFS~\cite{gupta2019lvis}$\star$} &  & 22.58 & 12.30 & 21.28 & 28.55 & 23.25 \\
 & & {\textcolor{cerise}{\cmark}} & \cellcolor{LightCyan}{(+2.65) 25.22} & \cellcolor{LightCyan}{(+7.03) 19.33} & \cellcolor{LightCyan}{(+2.88) 24.16} & \cellcolor{LightCyan}{(+0.43) 28.98} & \cellcolor{LightCyan}{(+2.83) 26.08} \\
 & \multirow{2}{*}{MosaicOS~\cite{zhang2021simple}} &  & 24.45 & 18.17 & 22.99 & 28.83 & 25.05 \\
 & & {\textcolor{cerise}{\cmark}} & \cellcolor{LightCyan}{(+2.32) {\textbf{26.76}}} & \cellcolor{LightCyan}{(+5.69) \textbf{23.86}}  & \cellcolor{LightCyan}{(+2.82) {25.82}} & \cellcolor{LightCyan}{(+0.27) {29.10}} & \cellcolor{LightCyan}{(+2.73) \textbf{27.77}} \\
 \midrule
 \multirow{7}{*}{R-101} & Seesaw~\cite{wang2020seesaw}$^{2\times}$ &  & 27.10 & 18.70 & 26.30 & 31.70 & 27.40 \\
 & Seesaw w/ RFS~\cite{wang2020seesaw}$^{2\times}$ &  & 28.10 & 20.00 & 28.00 & \textbf{31.90} & 28.90 \\
  \cmidrule(r){2-8}
& \multirow{2}{*}{RFS~\cite{gupta2019lvis}$\star$} &  & {24.82} & {15.18} & {23.71} & {30.31} & {25.45} \\
 & & {\textcolor{cerise}{\cmark}} & \cellcolor{LightCyan}{(+2.43) {27.25}} & \cellcolor{LightCyan}{(+5.61) {20.79}} & \cellcolor{LightCyan}{(+2.74) {26.45}} & \cellcolor{LightCyan}{(+0.68) {30.99}} & \cellcolor{LightCyan}{(+2.60) {28.05}} \\
 & \multirow{2}{*}{MosaicOS~\cite{zhang2021simple}} &  & {26.73} & {20.53} & {25.78} & {30.53} & {27.41} \\
 & & {\textcolor{cerise}{\cmark}} & \cellcolor{LightCyan}{(+2.30) \textbf{29.03}} & \cellcolor{LightCyan}{(+5.85) \textbf{26.38}} & \cellcolor{LightCyan}{(+2.37) \textbf{28.15}} & \cellcolor{LightCyan}{(+0.66) {31.19}} & \cellcolor{LightCyan}{(+2.55) \textbf{29.96}} \\
 \midrule
\multirow{6}{*}{X-101} & cRT~\cite{kang2019decoupling}$\clubsuit$ &  & 27.20 & 19.60 & 26.00 & 31.90 & -- \\
 & RIO~\cite{chang2021image} &  & 27.50 & 18.80 & 26.70 & 32.30 & 28.50 \\
  \cmidrule(r){2-8}
  & \multirow{2}{*}{RFS~\cite{gupta2019lvis}$\star$} &  & {26.67} & {17.60} & {25.58} & {31.89} & {27.35} \\
 & & {\textcolor{cerise}{\cmark}} & \cellcolor{LightCyan}{(+1.25) {27.92}} & \cellcolor{LightCyan}{(+2.15) {19.75}} & \cellcolor{LightCyan}{(+1.61) {27.19}} & \cellcolor{LightCyan}{(+0.45) {32.34}} & \cellcolor{LightCyan}{(+1.49) {28.83}} \\
 & \multirow{2}{*}{MosaicOS~\cite{zhang2021simple}} &  & {28.29} & {21.75} & {27.22} & {32.35} & {28.85} \\
 & & {\textcolor{cerise}{\cmark}} & \cellcolor{LightCyan}{(+1.52) \textbf{29.81}} & \cellcolor{LightCyan}{(+3.97) \textbf{25.72}} & \cellcolor{LightCyan}{(+1.70) \textbf{28.92}} & \cellcolor{LightCyan}{(+0.24) \textbf{32.59}} & \cellcolor{LightCyan}{(+1.71) \textbf{30.56}} \\
 \bottomrule
\end{tabular}
\end{table}

\subsection{Implementation}
\ourmethod is easy to implement and requires no re-training of the model. We follow Eq.~4 and Eq.~5 of the main paper to apply \ourmethod to the existing models. For all the baseline detectors, we directly take the released models from the corresponding papers without any modifications. We report the results on the validation set with the best hyper-parameter tuned on training images for all models and benchmarks. The implementations are mainly based on the Detectron2~\cite{wu2019detectron2} or MMdetection~\cite{mmdetection} framework. We run our experiments on 4 NVIDIA RTX A6000 GPUs with AMD 3960X CPUs.

\subsection{Inference and Evaluation}
We follow the standard evaluation protocol for the LVIS benchmark~\cite{gupta2019lvis}. Specifically, during the inference, the threshold of confidence score is set to $10^{-4}$, and we keep the top $300$ proposals as the predicted results. No test time augmentation is used. We adopt the standard mean Average Precision (AP) and denote the AP for rare, common, and frequent categories by $\text{AP}_{r}$, $\text{AP}_{c}$, and $\text{AP}_{f}$, respectively. For the object detection results on LVIS v0.5, we report the box AP for each category.


\section{Additional Experimental Results and Analyses}
\label{supp:s_results}

Due to space limitations, we only reported the results of \ourmethod with strong baseline models in the main paper (cf. Table 1). In this section, we provide detailed comparisons with more existing works on LVIS~\cite{gupta2019lvis} v1 and v0.5. We also examine \ourmethod on MSCOCO dataset~\cite{lin2014microsoft}. Moreover, we conduct further analyses and ablation studies of our method. 

\subsection{Results on LVIS v1 Instance Segmentation}
\label{supp:ss_v1_seg}

We summarize the results of instance segmentation on LVIS v1 in \autoref{tab:suppl_v1_seg}. As mentioned in \autoref{suppl:ss_impl_baseline}, several methods (\eg, BaGS~\cite{li2020overcoming}, EQL v2~\cite{tan2020equalizationv2}, Seesaw~\cite{wang2020seesaw}) change the decision/classification rules and it is not immediately obvious how to apply calibration to them. Nevertheless, we include their results for comparison. We observe, for example, that \ourmethod can improve a simple baseline such as RFS~\cite{gupta2019lvis} to match or outperform all methods but Seesaw~\cite{wang2020seesaw}, which is trained with a stronger 2$\times$ schedule and an improved mask head. When paired with MosaicOS~\cite{zhang2021simple}, \ourmethod can achieve state-of-the-art performance with all different backbone models, suggesting that improving the feature (especially on rare objects) and calibrating the classifier are key ingredients to the success of long-tailed object detection and instance segmentation.

\subsection{Results on LVIS v0.5 Instance Segmentation}
\label{supp:ss_v0.5_seg}

Many existing works focus on LVIS v0.5. In this subsection, we thus report the results of instance segmentation on LVIS v0.5 in \autoref{tab:suppl_v0.5_seg}. Again, we observe similar trends that \ourmethod can significantly improve the baseline models with all different backbone architectures. Particularly, we can also see improvements on the sigmoid-based object detector, \ie, BALMS~\cite{ren2020balanced}.

\subsection{Results on LVIS v0.5 Object Detection}
\label{supp:ss_v0.5_det}

In \autoref{tab:suppl_v0.5_det}, we compare with existing methods that reported results on LVIS v0.5 object detection --- only the bounding box annotations are used for model training. Concretely, we include EQL~\cite{tan2020equalization}, LST~\cite{hu2020learning}, BaGS~\cite{li2020overcoming}, TFA~\cite{wang2020frustratingly}, and MosaicOS~\cite{zhang2021simple}, as the compared methods. \newadd{In addition, we study a popular sigmoid-based detector, \ie, RetinaNet with focal loss~\cite{lin2017focal}. We train the RetinaNet using the default hyper-parameters~\cite{gupta2019lvis} and apply \ourmethod on top of it.} We see that \ourmethod can consistently improve the baseline models.
\begin{table}[h]
\centering
\tabcolsep 2pt
\fontsize{8}{9}\selectfont
\renewcommand\arraystretch{1}
\caption{\small \textbf{Instance segmentation results on the validation set of LVIS v0.5.} Our method \ourmethod can improve a simple baseline such as RFS~\cite{gupta2019lvis} to match or outperform all methods with different backbone models. {\color{red}{$\dagger$}}: slight performance drop on sigmoid-based detectors. $\star$: models from Detectron2~\cite{wu2019detectron2}. $\ddagger$: models from \cite{ren2020balanced} (the results are slightly different from those reported in \cite{ren2020balanced}).}
\label{tab:suppl_v0.5_seg}
\begin{tabular}{llcrrrrr}
\toprule
{Backbone} & {Method} & {\ourmethod} & {AP} & {\AP{r}} & {\AP{c}} & {\AP{f}} & {\APb} \\
\midrule
\multirow{17}{*}{R-50} & EQL~\cite{tan2020equalization} &  & 22.80 & 11.30 & 24.70 & 25.10 & 23.30 \\
 & LST~\cite{hu2020learning} &  & 23.00 & -- & -- & -- & -- \\
  & SimCal~\cite{wang2020devil} &  & 23.40 & 16.40 & 22.50 & 27.20 & -- \\
 & DropLoss~\cite{hsieh2021droploss} &  & 25.50 & 13.20 & 27.90 & 27.30 & 25.10 \\
 & Forest R-CNN~\cite{wu2020forest} &  & 25.60 & 18.30 & 26.40 & 27.60 & 25.90 \\
 & BaGS~\cite{li2020overcoming} &  & 26.25 & 17.97 & 26.91 & 28.74 & 25.76 \\
 & DisAlign~\cite{zhang2021distribution} &  & 24.20 & 8.50 & 26.20 & 28.00 & 23.90 \\
 & RIO~\cite{chang2021image} &  & 26.00 & 18.90 & 26.20 & 28.50 & -- \\
  & EQL v2~\cite{tan2020equalizationv2} &  & 27.10 & 18.60 & 27.60 & \textbf{29.90} & 27.00 \\
\cmidrule(r){2-8}
 & \multirow{2}{*}{BALMS~\cite{ren2020balanced}$\ddagger$} &  & {26.97} & {17.31} & {28.07} & {29.47} & {26.42} \\
 &  & {\textcolor{cerise}{\cmark}} & \cellcolor{LightCyan}{(+0.55) 27.52} & \cellcolor{LightCyan}{(+2.02) 19.33} & \cellcolor{LightCyan}{(+0.75) \textbf{28.82}} & \cellcolor{LightCyan}{{\color{red}{$\dagger$}}(-0.30) 29.17} & \cellcolor{LightCyan}{(+0.38) 26.80} \\
 & \multirow{2}{*}{RFS~\cite{gupta2019lvis}$\star$} &  & 24.39 & 15.98 & 23.97 & 28.26 & 23.64 \\
 &  & {\textcolor{cerise}{\cmark}} & \cellcolor{LightCyan}{(+2.23) 26.61} & \cellcolor{LightCyan}{(+2.73) 18.71} & \cellcolor{LightCyan}{(+3.40) 27.37} & \cellcolor{LightCyan}{(+0.57) 28.83} & \cellcolor{LightCyan}{(+2.36) 26.00} \\
 & \multirow{2}{*}{MosaicOS~\cite{zhang2021simple}} &  & {26.28} & {19.65} & {26.62} & {28.49} & {25.76} \\
 &  & {\textcolor{cerise}{\cmark}} & \cellcolor{LightCyan}{(+1.69) \textbf{27.97}} & \cellcolor{LightCyan}{(+3.57) \textbf{23.22}} & \cellcolor{LightCyan}{(+2.02) 28.64} & \cellcolor{LightCyan}{(+0.54) 29.03} & \cellcolor{LightCyan}{(+1.86) \textbf{27.61}} \\
 \midrule
\multirow{9}{*}{R-101} & EQL~\cite{tan2020equalization} &  & 26.20 & 11.90 & 27.80 & 29.80 & 26.20 \\
 & Forest R-CNN~\cite{wu2020forest} &  & 26.90 & 20.10 & 27.90 & 28.30 & 27.50 \\
 & DropLoss~\cite{hsieh2021droploss} &  & 26.90 & 14.80 & \textbf{29.80} & 28.30 & 26.80 \\
 & RIO~\cite{chang2021image} &  & 27.70 & 20.10 & 28.30 & 30.00 & 27.30 \\
 & EQL v2~\cite{tan2020equalizationv2} &  & 28.10 & \textbf{20.70} & 28.30 & \textbf{30.90} & \textbf{28.10} \\
 & DisAlign~\cite{zhang2021distribution}  &  & 25.80 & 10.30 & 27.60 & 29.60 & 25.60 \\
\cmidrule(r){2-8}
 & \multirow{2}{*}{RFS~\cite{gupta2019lvis}$\star$} &  & {25.75} & {15.46} & {25.96} & {29.60} & {25.44} \\
 &  & {\textcolor{cerise}{\cmark}} & \cellcolor{LightCyan}{(+2.38) \textbf{28.13}} & \cellcolor{LightCyan}{(+4.90) 20.36} & \cellcolor{LightCyan}{(+3.24) 29.20} & \cellcolor{LightCyan}{(+0.30) 29.90} & \cellcolor{LightCyan}{(+2.55) 28.00} \\
 \midrule
 \multirow{6}{*}{X-101} & Forest R-CNN~\cite{wu2020forest} &  & 28.50 & \textbf{21.60} & 29.70 & 29.70 & \textbf{28.80} \\
& RIO~\cite{chang2021image} &  & 28.90 & {19.50} & 29.70 & 31.60 & 28.60 \\
 & DisAlign~\cite{zhang2021distribution}  &  & 27.40 & 11.00 & 29.30 & 31.60 & 26.80 \\
\cmidrule(r){2-8}
 & \multirow{2}{*}{RFS~\cite{gupta2019lvis}$\star$} &  & {27.05} & {15.38} & {27.34} & {31.35} & {26.66} \\
 &  & {\textcolor{cerise}{\cmark}} & \cellcolor{LightCyan}{(+1.93) \textbf{28.98}} & \cellcolor{LightCyan}{(+3.94) 19.32} & \cellcolor{LightCyan}{(+2.60) \textbf{29.94}} & \cellcolor{LightCyan}{(+0.27) \textbf{31.62}} & \cellcolor{LightCyan}{(+1.94) 28.60} \\
 \bottomrule
\end{tabular}
\end{table}

\begin{table}[h]
\centering
\tabcolsep 6pt
\fontsize{8}{9}\selectfont
\renewcommand\arraystretch{1}
\caption{\small \textbf{Object detection results on the validation set of LVIS v0.5.} \ourmethod significantly boosts baseline methods. All models use Faster R-CNN~\cite{ren2016faster} with ResNet-50 and FPN~\cite{lin2017feature}. {\color{red}{$\dagger$}}: slight drop on frequent class. $\clubsuit$: pre-trained with MSCOCO~\cite{lin2014microsoft}. \newadd{$\S$: models trained by ourselves.} $\star$: models from~\cite{zhang2021simple}. $\ddagger$: models from \cite{li2020overcoming}.}
\label{tab:suppl_v0.5_det}
\begin{tabular}{lcrrrr}
\toprule
Method & \ourmethod & $\text{AP}^{b}$ & $\text{AP}^{b}_{r}$ & $\text{AP}^{b}_{c}$ & $\text{AP}^{b}_{f}$ \\
\midrule
EQL~\cite{tan2020equalization} &  & 23.30 & -- & -- & -- \\
LST~\cite{hu2020learning} &  & 22.60 & -- & -- & -- \\
BaGS~\cite{li2020overcoming}$\clubsuit$ &  & 25.96 & 17.66 & 25.75 & 29.55 \\
\midrule
\multirow{2}{*}{RetinaNet~\cite{lin2017focal}$\S$} &  & 16.34 & 9.47 & 14.07 & 21.93 \\
 & {\textcolor{cerise}{\cmark}} & \cellcolor{LightCyan}{(+0.98) 17.32} &  \cellcolor{LightCyan}{(+2.24) 11.71} & \cellcolor{LightCyan}{(+1.62) 15.69} & \cellcolor{LightCyan}{{\color{red}{$\dagger$}}(-0.32) 21.61} \\

\multirow{2}{*}{Faster R-CNN~\cite{ren2016faster}$\clubsuit,\ddagger$} &  & 20.98 & 4.13 & 19.70 & 29.30 \\
 & {\textcolor{cerise}{\cmark}} & \cellcolor{LightCyan}{(+2.89) 23.87} &  \cellcolor{LightCyan}{(+2.85) 6.98} & \cellcolor{LightCyan}{(+4.47) 24.17} & \cellcolor{LightCyan}{(+0.94) 30.24} \\
\multirow{2}{*}{RFS~\cite{gupta2019lvis}$\star$} &  & 23.35 & 12.98 & 22.60 & 28.42 \\
 & {\textcolor{cerise}{\cmark}} & \cellcolor{LightCyan}{(+2.27) 25.62} & \cellcolor{LightCyan}{(+4.57) 17.55} & \cellcolor{LightCyan}{(+2.93) 25.53} & \cellcolor{LightCyan}{(+0.53) 28.95} \\
\multirow{2}{*}{TFA~\cite{wang2020frustratingly}} &  & 24.07 & 14.90 & 23.89 & 27.94 \\
 & {\textcolor{cerise}{\cmark}} & \cellcolor{LightCyan}{(+0.56) 24.63} & \cellcolor{LightCyan}{(+1.72) 16.62} & \cellcolor{LightCyan}{(+0.84) 24.73} & \cellcolor{LightCyan}{{\color{red}{$\dagger$}}(-0.25) 27.70} \\
\multirow{2}{*}{MosaicOS~\cite{zhang2021simple}} &  & 25.01 & 20.19 & 23.89 & 28.33 \\
 & {\textcolor{cerise}{\cmark}} & \cellcolor{LightCyan}{(+2.53) 27.54} & \cellcolor{LightCyan}{(+4.88) \textbf{25.07}} & \cellcolor{LightCyan}{(+3.32) 27.21} & \cellcolor{LightCyan}{(+0.60) {28.93}} \\
\multirow{2}{*}{MosaicOS~\cite{zhang2021simple}$\clubsuit$} &  & 26.30 & 17.32 & 26.20 & 30.00 \\
 & {\textcolor{cerise}{\cmark}} & \cellcolor{LightCyan}{(+2.05) \textbf{28.35}} & \cellcolor{LightCyan}{(+5.82) 23.14} & \cellcolor{LightCyan}{(+2.19) \textbf{28.39}} & \cellcolor{LightCyan}{(+0.37) \textbf{30.37}} \\
 \bottomrule
\end{tabular}
\end{table}

\newadd{
\subsection{Results on Objects365 dataset}
\label{supp:ss_object365}

We further validate \ourmethod on Objects365~\cite{shao2019objects365}, a dataset designed to spur object detection research with a focus on diverse objects in the wild. Objects365 contains 2 million images, 30 million bounding boxes, and 365 categories with a long-tailed distribution. We train a Faster R-CNN~\cite{ren2016faster} as the baseline on the training set, with FPN and ResNet-50 as the backbone. We report results in \autoref{tab:suppl_objects365}. We not only show the overall mean AP, but also the mean APs for different groups of categories based on the training image number per category. \ourmethod outperforms the baseline detector on all groups of categories, justifying its effectiveness and generalizability.
}

\begin{table}[h]
\centering
\tabcolsep 6pt
\fontsize{8}{9}\selectfont
\renewcommand\arraystretch{1.0}
\caption{\small \newadd{{Results of object detection AP within each group of categories (according to the training image numbers) on Objects365~\cite{shao2019objects365} validation set. The baseline model is Faster R-CNN with ResNet-50 and FPN.}}}
\label{tab:suppl_objects365}
\begin{tabular}{lrrrrrr}
\toprule
 & AP & $\text{AP}_{(0, 100)}$ & $\text{AP}_{[100, 1000)}$ & $\text{AP}_{[1000, 10000)}$ & $\text{AP}_{[10000, +\infty)}$ \\
\midrule
\# Category & 365 & 33 & 115 & 141 & 76 \\
\midrule
Baseline & 16.29 &	2.43 &	6.95 &	20.88 &	27.93 \\
w/ \ourmethod & (+0.48) 16.77 &	(+0.23) 2.67 & 	(+0.54) 7.50 &	(+0.48) 21.36 &	(+0.50) 28.43 \\
\bottomrule
\end{tabular}
\end{table}

\subsection{Results on MSCOCO Dataset}
\label{supp:ss_mscoco}

\begin{figure}[h]
    \centering
    \includegraphics[width=1\textwidth]{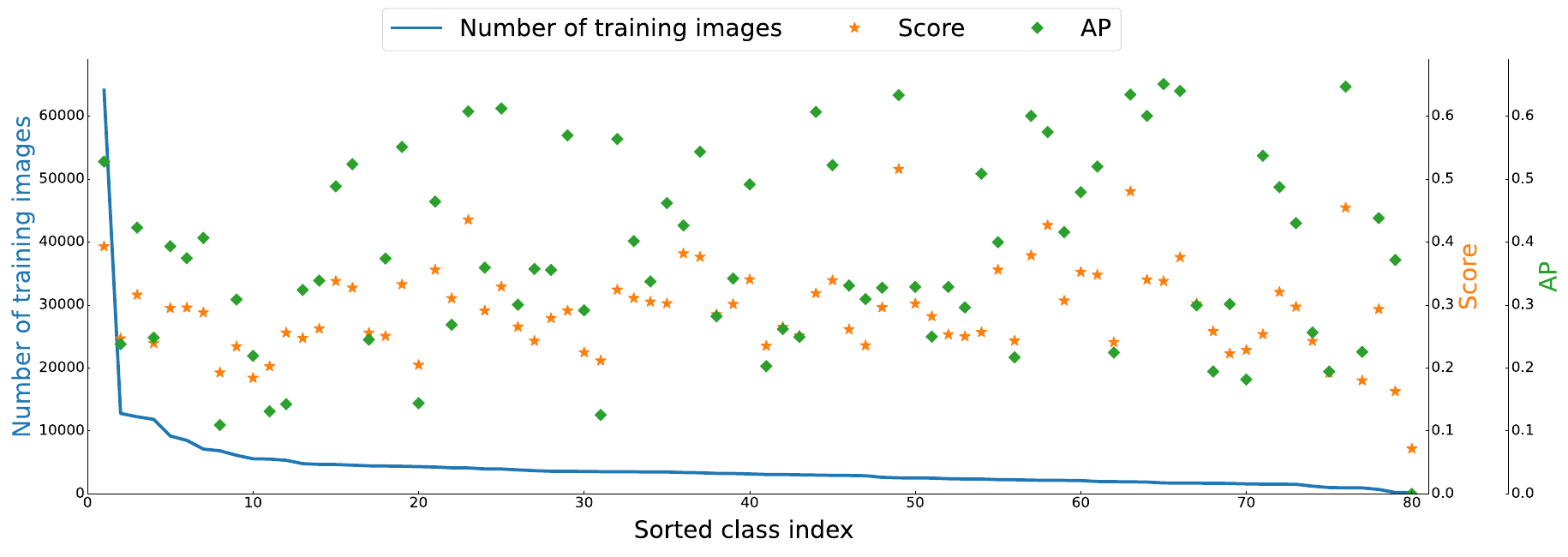}
    \caption{\small \textbf{Per-class AP of Faster R-CNN and the category distribution on MSCOCO (2017).} The categories are sorted in descending {\color{my_b}{numbers of training images}}. 
   {\color{my_a}{Orange stars}} indicate the average of predicted confidence scores for each class. {\color{my_c}{Green diamonds}} are per-class APs. The least frequent class, “hair driver”, still has 189 training images, indicating that all the classes in MSCOCO are considered as frequent classes using the definition of LVIS.}
    \label{fig:suppl_mscoco}
\end{figure}

We also experiment our method \ourmethod on the generic object detection benchmark, \ie, MSCOCO~\cite{lin2014microsoft}. MSCOCO is the most popular benchmark for object detection and instance segmentation, which contains 80 categories with a relative balanced class distribution (See \autoref{fig:suppl_mscoco}). More importantly, the least frequent class, ``hair driver'', still has $189$ training images. In other words, all the classes in MSCOCO are considered as frequent classes using the definition of LVIS.
We report results in \autoref{tab:suppl_mscoco}. We see that the performance gains brought by \ourmethod is marginal. Our hypothesis is that the detectors trained with MSCOCO already see sufficient examples for all categories (even for tail classes) and the trained classifier is less biased.  

\begin{table}[h]
\centering
\tabcolsep 6pt
\fontsize{8}{9}\selectfont
\renewcommand\arraystretch{1.0}
\caption{\small \textbf{Results of object detection on MSCOCO~\cite{lin2014microsoft}.} The baseline model is from Faster R-CNN with FPN and ResNet-50 as the backbone.}
\label{tab:suppl_mscoco}
\begin{tabular}{lcccccc}
\toprule
Method & AP & $\text{AP}_{50}$ & $\text{AP}_{75}$ & $\text{AP}_{s}$ & $\text{AP}_{m}$ & $\text{AP}_{l}$ \\
\midrule
Baseline & 37.93 & 58.84 & 41.05 & 22.44 & 41.14 & 49.10 \\
w/ \ourmethod & 37.96 & 58.40 & 41.22 & 22.22 & 41.18 & 49.48 \\
\bottomrule
\end{tabular}
\end{table}

\newadd{
\subsection{Results on Image Classification Datasets}
\label{supp:ss_classification}

Besides object detection and instance segmentation, we further evaluate \ourmethod on three imbalanced classification benchmarks: ImageNet-LT~\cite{liu2019large}, iNaturalist (2018 version)~\cite{van2018inaturalist}, and CIFAR-10-LT (with an imbalance factor 100)~\cite{CBfocal}. ImageNet-LT has 1,000 classes while iNaturalist has 8,142 classes. All three datasets have long-tailed distributions on the number of training images per class but have a balanced evaluation set. We follow the literature to train a ResNet-50 classifier for the first two datasets, and a ResNet-32 classifier for CIFAR. Since there is no background class in these datasets, we simply drop the background class in Eq.4 in the main text. Results are shown in \autoref{tab:suppl_classification}. As expected, \ourmethod consistently outperforms the baseline classifiers, demonstrating its effectiveness on long-tailed classification problems as well.

As mentioned in the Section 1 in the main paper, post-processing calibration for imbalanced or long-tailed classification has been studied in several prior works. Our approach is indeed inspired by their efficiency and effectiveness in classification problems and we extend them to the detection and instance segmentation problems.
}

\begin{table}[h]
\tabcolsep 9pt
\centering
\fontsize{8}{9}\selectfont
    \caption{\small \newadd{\textbf{Classification accuracy} on ImageNet-LT~\cite{liu2019large}, iNaturalist~\cite{van2018inaturalist}, and CIFAR-10-LT~\cite{CBfocal}.}}    
    \label{tab:suppl_classification}
    \begin{tabular}{l r  r}
    \multicolumn{3}{c}{{(a) ImageNet-LT}}\\
    \toprule
      Method & Top-1 &  Top-5 \\
      \midrule
      Baseline  & 45.11 &  71.18\\
      w/ \ourmethod  &  49.71  &  74.53 \\
      \bottomrule
    \end{tabular}
    \hfill
    \begin{tabular}{l r r}
    \multicolumn{3}{c}{{(b) iNaturalist}}\\
      \toprule
      Method & Top-1 &  Top-5 \\
      \midrule
      Baseline  & 61.54 & 82.94 \\
      w/ \ourmethod  &  65.15 &  84.83\\
      \bottomrule
    \end{tabular}
    \hfill
    \begin{tabular}{l r}
    \multicolumn{2}{c}{{(c) CIFAR-10-LT}}\\
      \toprule
      Method & Top-1  \\
      \midrule
      Baseline  &  70.36 \\
      w/ \ourmethod  &  77.78 \\
      \bottomrule
    \end{tabular}
\end{table}

\subsection{Ablation Studies on Sigmoid-Based Detectors (\ie, with Multiple Binary Classifiers)}
\label{supp:ss_ablation}

As shown in the main paper (cf. Table 3), we conduct ablation studies of \ourmethod with a standard softmax-based object detection. Here, we further examine a sigmoid-based object detector, \ie, BALMS~\cite{ren2020balanced}, and report the results in \autoref{tab:suppl_ablation}. Beyond Eq.~7 of the main paper, we ablate \ourmethod with different calibration mechanisms, factors, and with and without score normalization.
We note that, in this kind of models, $C$ binary classifiers are learned, each corresponds to one foreground class. In other words, no background class is specifically learned. Thus, the score normalization is usually not necessary or harmful --- the background patches with low scores by all the classifiers will now gets their scores boosted due to calibration.
\begin{table}[th]
\centering
        \fontsize{8}{9}\selectfont
        \tabcolsep 8pt
        \renewcommand\arraystretch{1.0}
        \centering
        \caption{\small 
            \textbf{Ablation studies of \ourmethodbold with the sigmoid-based baseline model (BALMS~\cite{ren2020balanced}).} We follow \citet{ren2020balanced} to report the results on LVIS v0.5 instance segmentation. \cali: calibration mechanism. \nor: class score normalization. The best ones are in bold. As discussed in \autoref{supp:ss_ablation}, normalization is not suitable for this kind of models.
        }
        \label{tab:suppl_ablation}
        \begin{tabular}{ccccccc}
        \toprule
        {$a_c$} & {\cali} & {\nor} & {AP} & {\AP{r}} & {\AP{c}} & {\AP{f}} \\
        \midrule
        \multicolumn{1}{c}{Baseline} &  $\exp(-\phi_c(\vx))$ &  {\textcolor{cerise}{\cmark}} & 26.97&	17.31&	28.07&	\textbf{29.47}\\
        \midrule
        \multirow{7}{*}{\makecell[c]{$\cfrac{1 - \gamma^{N_c}}{1 - \gamma}$ \\ (ENS~\cite{CBfocal})}} & \multirow{2}{*}{$\exp(-\phi_c(\vx) \times a_c)$} & {{\textcolor{codegray}{\xmark}}} & 26.99&	17.40&	28.06&	29.46 \\
         &  & {\textcolor{cerise}{\cmark}} & 15.56&	7.73&	14.91&	19.51 \\
        \cmidrule(r){2-7}
         & \multirow{2}{*}{$s_c \times a_c$} & {{\textcolor{codegray}{\xmark}}} & 27.12	 &\textbf{19.89}	 &28.25 &	28.59 \\
         &  & {\textcolor{cerise}{\cmark}} & 15.29&	12.05&	16.98&	14.47\\
         \cmidrule(r){2-7}
         & \multirow{2}{*}{$\exp(-\phi_c(\vx)) \times a_c$} & {{\textcolor{codegray}{\xmark}}} & 27.17&	19.88&	28.26&	28.71\\
         &  & {\textcolor{cerise}{\cmark}} & 18.62&	12.34&	18.29&	21.55 \\
         \midrule
         \multirow{7}{*}{\makecell[c]{$N_{c}^{\gamma}$\\(CDT~\cite{ye2020identifying})}} & \multirow{2}{*}{$\exp(-\phi_c(\vx) \times a_c)$} & {\textcolor{codegray}{\xmark}} & 27.37&	18.64&	28.69&	29.22  \\
         &  & {\textcolor{cerise}{\cmark}}  & 16.82&	9.50&	17.24&	19.22 \\
        \cmidrule(r){2-7}
         & \multirow{2}{*}{$s_c \times a_c$} & {{\textcolor{codegray}{\xmark}}} & 27.52 &	19.33 &	\textbf{28.82} &	29.17 \\
         &  & {\textcolor{cerise}{\cmark}} & 15.62&	11.58&	17.32&	15.10 \\
         \cmidrule(r){2-7}
         & \multirow{2}{*}{$\exp(-\phi_c(\vx)) \times a_c$} & {{\textcolor{codegray}{\xmark}}} & \cellcolor{LightCyan}{\textbf{27.52}}&	\cellcolor{LightCyan}{19.34}&	\cellcolor{LightCyan}{28.80}&	\cellcolor{LightCyan}{29.19} \\
         &  & {\textcolor{cerise}{\cmark}} & 18.60&	12.64&	18.36&	21.28 \\
         \bottomrule
        \end{tabular}
\end{table}

\subsection{Empirical Class Frequency is Better than Classifier Norms for \ourmethodbold}
\label{supp:ss_weight_norm}
As mentioned in the main paper (cf. Section 3.3 and Table 2 (bottom)), class-dependent temperature ($N_c^\gamma$)~\cite{ye2020identifying} provides a better signal for calibration than the classifier norms ($\|\vw_c\|_2^{\gamma}$) of the classifier. \autoref{tab:suppl_comp_weight_norm} shows a comparison between those two factors for our proposed calibration mechanism. With \ourmethod, we see that $N_c$ outperforms $\|\vw_c\|_2$ on all object categories. Moreover, we notice that leaving the background intact shows a better performance, justifying our analysis and experimental results on how to handle the background class (cf. Section 3.2 and Figure 4 of the main paper).

\begin{table}[th]
\centering
\tabcolsep 6pt
\fontsize{8}{9}\selectfont
\renewcommand\arraystretch{1.0}
\caption{\small \textbf{Empirical class frequency ($N_c$) is better than classifier norms ($\|\vw_c\|_2$) for \ourmethodbold.} Results are reported on LVIS v1 instance segmentation. Background: whether calibrating the background class or not.}
\label{tab:suppl_comp_weight_norm}
\begin{tabular}{cccccccc}
\toprule
Method & $a_c$ & Background & {AP} & {\AP{r}} & {\AP{c}} & {\AP{f}} & {\APb} \\
\midrule
RFS~\cite{gupta2019lvis} & -- & -- & 22.58 & 12.30 & 21.28 & 28.55 & 23.25 \\
\midrule
\multirow{3}{*}{w/ \ourmethod} & \multirow{2}{*}{$\|\vw_c\|^\gamma_2$} & \xmark & 22.86 & 13.21 & 21.67 & 28.43 & 23.41 \\
 &  & \cmark & 22.56 & 12.47 & 21.34 & 28.37 & 23.17 \\
 \cmidrule(r){2-8}
 & $N_c^\gamma$ & \xmark & 25.22 & 19.33 & 24.16 & 28.98 & 26.08 \\
 \bottomrule
\end{tabular}
\end{table}

\subsection{Further Analysis on Existing Post-Processing Calibration Methods}
\label{supp:ss_post_processing}

We compare \ourmethod to the existing post-calibration methods in the main paper (cf. Table 2 (upper)). In the main paper, we follow the implementations in \cite{dave2021evaluating} to perform the calibration after the top $300$ predicted boxes are selected. Here we study an alternative of directly applying the calibration before selecting the $300$ predictions. We show the results in \autoref{tab:suppl_post_processing}. \ourmethod still outperforms all existing calibration methods.

\begin{table}[th]
\centering
\tabcolsep 8pt
\fontsize{8}{9}\selectfont
\renewcommand\arraystretch{1.0}
\caption{\small \textbf{Further analysis and comparison on existing post-processing calibration methods.} Results are reported on LVIS v1 instance segmentation. When to calibrate: before or after the top $300$ predicted boxes are selected per image.
}
\label{tab:suppl_post_processing}
\begin{tabular}{lccccc}
\toprule
Method & \multicolumn{1}{c}{When to calibrate?} & AP & APr & APc & APf \\
\midrule
RFS~\cite{gupta2019lvis} & -- & 22.58 & 12.30 & 21.28 & 28.55 \\
\midrule
\multirow{2}{*}{w/ HistBin~\cite{zadrozny2001obtaining}} & before & 18.91 & 5.65 & 17.49 & 26.33 \\
 & after & 21.82 & 11.28 & 20.31 & 28.13 \\
 \midrule
\multirow{2}{*}{w/ BBQ (AIC)~\cite{naeini2015obtaining}} & before & 16.56 & 3.07 & 14.76 & 24.51 \\
 & after & 22.05 & 11.41 & 20.72 & 28.21 \\
  \midrule
\multirow{2}{*}{w/ Beta calibration~\cite{kull2017beta}} & before & 22.11 & 11.54 & 21.77 & 27.15 \\
 & after & 22.55 & 12.29 & 21.27 & 28.49 \\
  \midrule
\multirow{2}{*}{w/ Isotonic seg.~\cite{zadrozny2002transforming}} & before & 20.58 & 10.46 & 20.36 & 25.27 \\
 & after & 22.43 & 12.19 & 21.12 & 28.41 \\
  \midrule
\multirow{2}{*}{w/ Platt. scaling~\cite{platt1999probabilistic}} & before & 22.09 & 12.07 & 21.40 & 27.26 \\
 & after & 22.55 & 12.29 & 21.27 & 28.49 \\
  \midrule
w/ \ourmethod & before & 25.22 & 19.33 & 24.16 & 28.98 \\
\bottomrule
\end{tabular}
\vspace{-3mm}
\end{table}

\subsection{Additional Qualitative Results}
\label{supp:ss_qual}

\begin{figure}[t]
    \centering
    \includegraphics[width=0.76\textwidth]{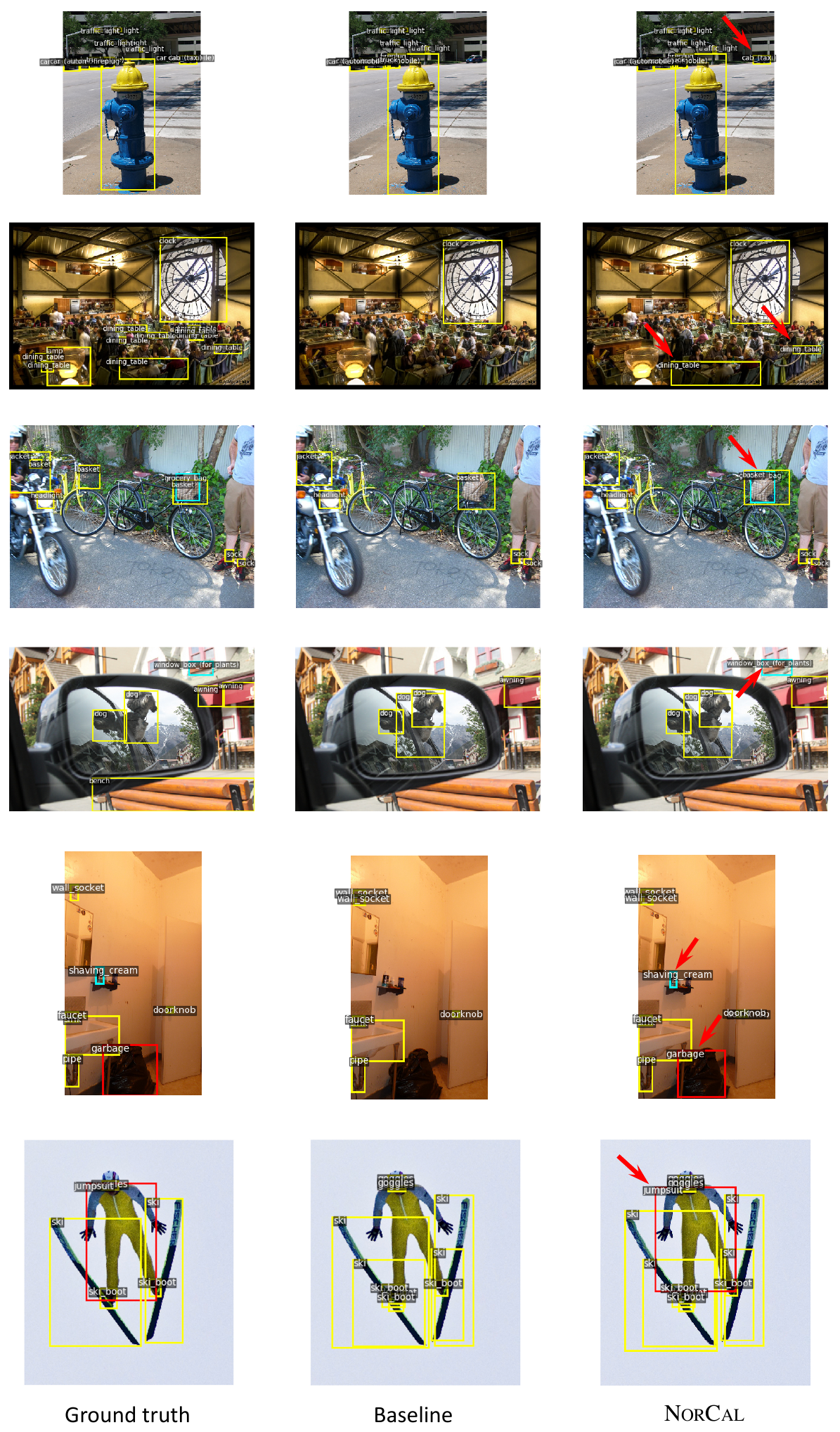}
    \caption{\small \textbf{Additional qualitative results.} We superimpose {\color{red}{red}} arrows to show the improvement.  
    {\color{yellow}{Yellow}}, {\color{cyan}{cyan}} and {\color{red}{red}} bounding boxes indicate frequent, common and rare class labels.}
    \label{fig:suppl_qual}
\end{figure}

We provide additional qualitative results on LVIS v1 in \autoref{fig:suppl_qual}. We show the (predicted) bounding boxes from the ground truth annotations, the baseline Mask R-CNN~\cite{he2017mask} with RFS~\cite{gupta2019lvis}, and \ourmethod.

\end{document}